\documentclass[12pt]{article}%
\usepackage{amsmath}
\usepackage{amsfonts}
\usepackage{amssymb}
\usepackage{longtable}
\usepackage{multirow}
\usepackage{rotating}
\usepackage{graphicx}%
\usepackage{color}
\usepackage{graphics}
\usepackage{graphicx}
\usepackage{epsfig}
\usepackage{subfigure}
\usepackage{latexsym,bm}
\usepackage{graphicx}%
\usepackage{mathrsfs}
\usepackage{cite}
\usepackage{mathtools}
\usepackage{url}
\usepackage{hyperref}
\usepackage{diagbox} 
\usepackage{pifont}

\usepackage{algorithm}
\usepackage{algpseudocode}

\usepackage[usenames,dvipsnames]{xcolor}
\usepackage{tcolorbox}
\usepackage{tabularx}
\usepackage{array}
\usepackage{colortbl}
\tcbuselibrary{skins}

\usepackage[T1]{fontenc}
\usepackage[utf8]{inputenc}

\newtheorem{example}{Example}

\begin{document}

\def\R{{\mathbb R}}
\newcommand{\bbb}[1]{{\boldsymbol  #1 }}
\newcommand{\vth}{{\vartheta}}
\newcommand{\blue}[1]{{\color{blue} #1 }}
\newcommand{\red}[1]{{\color{red} #1 }}
\newcommand{\magenta}[1]{{\color{magenta} #1 }}
\newcommand{\g}[1]{{\color{teal} #1 }}

\newcommand{\eff}{$\kappa_{\rm eff}$}
\title{Highway Networks for Improved Surface Reconstruction: The Role of Residuals and Weight Updates}


\author{
A. Noorizadegan\footnote{Department of Civil Engineering, National Taiwan University, 10617, Taipei, Taiwan}
, Y.C. Hon\footnote{Department of Mathematics, Chinese University of Hong Kong, SAR, Hong Kong, China}
, D.L. Young\footnote{Core Tech System Co. Ltd, Moldex3D, Chubei, Taiwan} \footnotemark[1] 
, C.S. Chen\footnotemark[1] \footnotemark[3] 
\footnote{Corresponding author: dchen@ntu.edu.tw }
}

\date{}

\maketitle

\begin{abstract}

Surface reconstruction from point clouds is a fundamental challenge in computer graphics and medical imaging. In this paper, we explore the application of
advanced neural network architectures for the accurate and efficient reconstruction of surfaces from data points. We introduce a novel variant of the Highway network (Hw) called Square-Highway (SqrHw) within the context of multilayer perceptrons and investigate its performance alongside plain neural networks and a simplified Hw in various numerical
examples. These examples include the reconstruction of simple and complex surfaces, such as spheres, human hands, and intricate models like the Stanford Bunny.
We analyze the impact of factors such as the number of hidden layers, interior and
exterior points, and data distribution on surface reconstruction quality. Our results
show that the proposed SqrHw architecture outperforms other neural network
configurations, achieving faster convergence and higher-quality surface reconstructions. Additionally, we demonstrate the SqrHw’s ability to predict surfaces
over missing data, a valuable feature for challenging applications like medical imaging. Furthermore, our study delves into further details, demonstrating that the proposed method based on highway networks yields more stable weight norms and backpropagation gradients compared to the Plain Network architecture. This research not only advances the field of computer graphics but also holds utility for other purposes such as function interpolation and physics-informed neural networks, which integrate multilayer perceptrons into their algorithms. The codes implemented are available at: 
\url{https://github.com/CMMAi/ResNet_for_PINN}

\end{abstract}

\noindent{Keywords: Surface reconstruction, surface rendering, point clouds,
deep learning, computer graphics, medical imaging, highway network, physics-informed neural networks. }

\section{Introduction}
\label{sec:intro}
Computer graphics is a rapidly evolving field that continually seeks innovative techniques for surface reconstruction and visualization, offering a broad spectrum of applications from 3D object modeling to animation, rendering, and beyond \cite{Erler20}.

Surface reconstruction from point clouds has been a subject of considerable research effort, where methods are categorized into those that employ \textbf{data-driven with prior knowledge} and \textbf{non-data-driven}. Data-driven with prior knowledge methods (the first class) have emerged as potent solutions, deriving priors from extensive datasets. These methods, including AtlasNet \cite{Groueix18}, Scan2Mesh \cite{Dai19}, Points2surf \cite{Erler20}, and others, leverage feature representations to adeptly handle noisy or partial input. While these algorithms necessitate extensive datasets and a latent feature space, with a primary focus on encoding shapes that fall within this feature space, our approach charts a distinctive course. 

On the other hand, non-data-driven class including methods like method of fundamental  solutions \cite{Zheng20}, radial basis function method \cite{Liu20}, scale space meshing \cite{Digne11} and Ohrhallinger's combinatorial approach \cite{Ohrhallinger13}, offers effective solutions but often struggles with partially  data or non-smooth surfaces. Additionally, deformations and patch-based methods \cite{Li10,Williams19} from this class prove to be limited in handling complex topologies and connectivity changes. Poisson reconstruction \cite{Kazhdan06,Kazhdan13} stands as the prevailing benchmark in \textit{non-data-driven} surface reconstruction from point clouds. Notably, none of the previously mentioned methods employ a prior that distills information about typical surface shapes from a vast dataset.

While the classical methods used in non-data-driven class are rooted in rigorous mathematics, difficulties arise in selecting uncertainties, such as the shape parameter in RBF methods or the location of center points in the method of fundamental solutions, making them challenging. Additionally, these methods often rely on partial differential equations (PDEs) like the Poisson equation \cite{Tankelevich09} or modified Helmholtz equation \cite{Liu20}, which further complicate implementation. Liu et al. \cite{Liu20} demonstrated significant sensitivity of results to parameters such as the shape parameter and $\lambda$ (a penalty factor in modified Helmholtz PDE), as illustrated in Figure 7 of their work, highlighting the challenges in uncertainty management within the method of fundamental solutions.

Our approach falls under the category of non-data-driven methods, employing deep learning methods in particular feedforward (multi-layer perceptrons) architecture. In contrast to traditional non-data-driven methods that solve partial differential equations, our method focuses on simple function interpolation (approximation). While easier, deep learning methods can face stability and accuracy challenges. Drawing inspiration from advancements such as highway networks \cite{Srivastava15a,Srivastava15} and subsequent residual networks (ResNet) \cite{He15,He16}, which enhance stability and accuracy in networks like convolutional and multi-layer perceptrons architectures, we introduce two novel networks. Our residual-based architecture significantly enhances both accuracy and stability.

These simplified residual-based architectures hold promise across various surface reconstruction scenarios and have shown efficacy in recent studies addressing interpolation and inverse problems \cite{Amir23,Amir23a}.

Our research objectives encompass:
\begin{itemize}
  \item A comprehensive analysis of neural network architectures for surface reconstruction from point clouds within computer graphics.
    \begin{itemize}
      \item Our proposed residual-based networks outperform conventional plain neural networks.
      \item They exhibit faster convergence and more stable optimization behavior due to the integration of a carry gate.
      \item Empirical evaluations across diverse datasets confirm their effectiveness in accurately reconstructing surfaces.
      \item Analysis of network parameters and hyperparameters underscores the importance of selecting appropriate architecture configurations.
    \end{itemize}
  \item Examination of weight updates reveals that residual-based networks converge to stable weight norms with smoother behavior and smaller norms compared to plain networks, indicative of superior optimization performance.
  \item Study of back-propagated gradients highlights the method's ability to mitigate the vanishing gradient issue, resulting in enhanced training stability and accelerated convergence rates.
\end{itemize}

\noindent
Our comparisons show that our proposed architecture has significant advantages for surface reconstruction in computer graphics. 
In Section 2, we introduce plain neural networks and their basic methodology. Section 3 presents the algorithm for using plain networks in computer graphics problems. Section 4 discusses advanced network architectures, including the proposed networks. In Section 5, we demonstrate the strengths of the proposed methods through various computer graphics examples. Finally, we conclude our findings in Section 6.

\section{Plain Neural Networks (Pn)}
\subsection{Multi-Layer Perceptrons (MLPs)}

In this section, we will explore the intricate architecture and operational dynamics of Multi-Layer Perceptrons (MLPs), which form the bedrock of numerous deep learning frameworks. MLPs, represented by the symbol $\mathcal{M}$, are meticulously crafted to approximate a function $\textbf{f} : \textbf{p} \in \mathbb{R}^d \rightarrow \textbf{y} \in \mathbb{R}^D$ through the strategic arrangement of artificial neurons across successive layers.

\subsection{Layer Configuration}
The typical composition of an MLP includes:

\begin{itemize}
    \item \textbf{Input Layer}: Receives input data of dimension \(d\).
    \item \textbf{Hidden Layers}: Perform computations to uncover complex patterns.
    \item \textbf{Output Layer}: Produces the network's predictions with dimensionality \(D\).
\end{itemize}

\begin{figure}
\centering%
\includegraphics[width=4.3in]{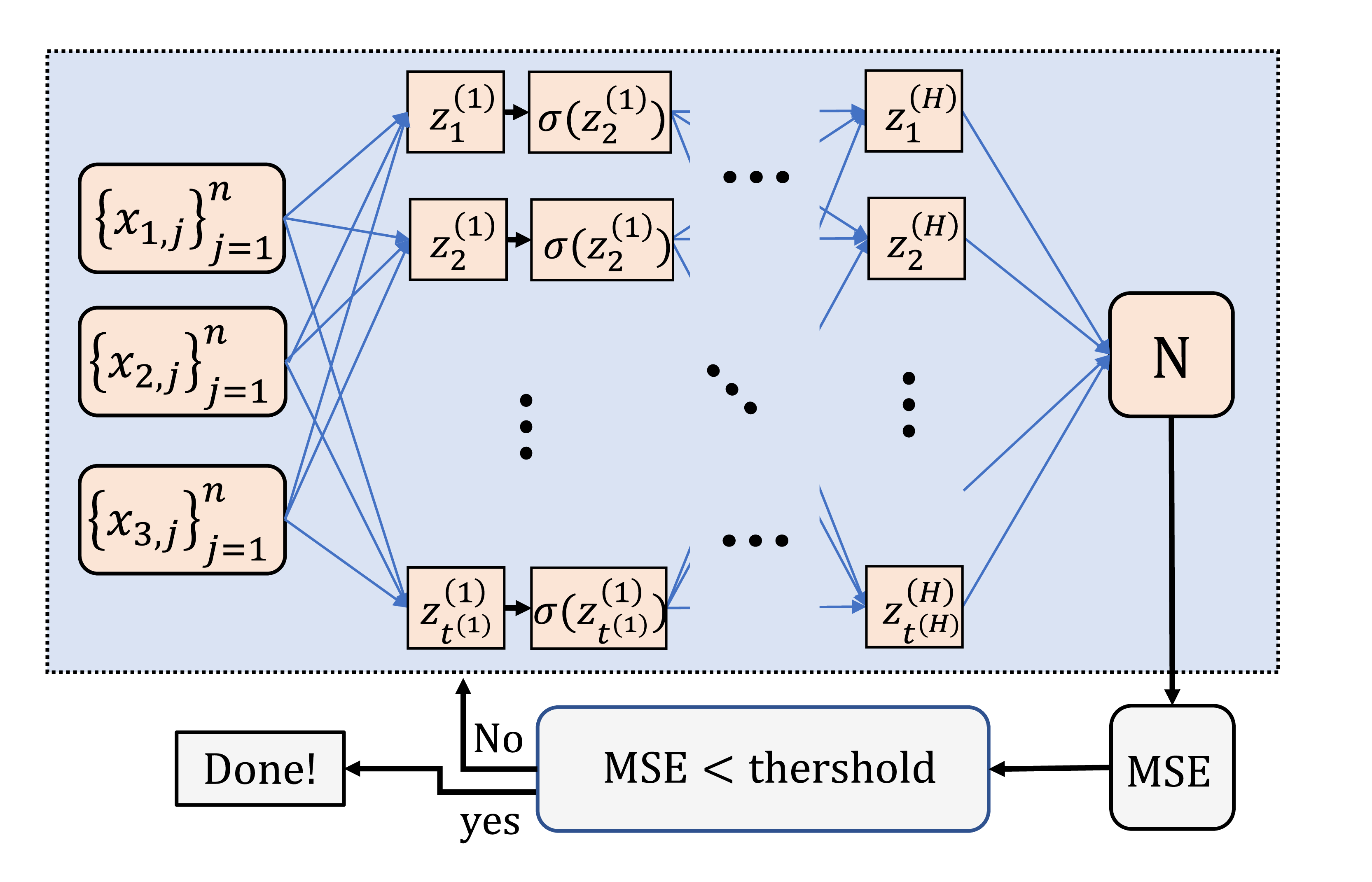}
\caption{schematic of an MLP.  } \label{MLP}
\end{figure}

The width of each layer, denoted as \(t^{(h)}\), dictates the number of neurons within that layer. Let's consider a network with \(H\) hidden layers, where the output vector for the \(h\)-th layer is denoted as \(\textbf{p}^{(h)} \in \mathbb{R}^{t^{(h)}}\), serving as the input to the subsequent layer. The input signal provided by the input layer is denoted as \(\textbf{p}^{(0)} = \textbf{p} \in \mathbb{R}^d\).

In each layer \(h\), \(1 \leq h \leq H + 1\), the \(i\)-th neuron performs an affine transformation followed by a non-linear operation:

\begin{equation}\label{eq1}
    z^{(h)}_i = W^{(h)}_{ij} {\rm{p}}^{(h-1)}_j + b^{(h)}_i,
\end{equation}
where \(1 \leq i \leq t^{(h)}\) and \( 1 \leq j \leq t^{(h-1)}\). Thus,

\begin{equation}\label{eq2}
    {\rm{p}}^{(h)}_i = \sigma(z^{(h)}_i),
\end{equation}
for \(1 \leq i \leq t^{(h)}\). Here, \(W^{(h)}_{ij}\) and \(b^{(h)}_i\) represent the weights and biases associated with the \(i\)-th neuron of layer \(h\), respectively, while \(\sigma(\cdot)\) denotes the activation function, which, in our case, is \(\tanh\).
The overall behavior of the network, denoted as \(\mathcal{M} : \mathbb{R}^d \rightarrow \mathbb{R}^D\), can be conceptually understood as a sequence of alternating affine transformations and component-wise activations, as depicted in Eqs. \eqref{eq1}-\eqref{eq2}. The architecture of a Multilayer Perceptron is illustrated schematically in Fig. \ref{MLP}, where \(x_1\), \(x_2\), and \(x_3\) represent three dimensions (\(\textbf{p} = (x_1, x_2, x_3)\)), each containing \(n\) samples. In this figure, \(\textbf{N}\) represents the approximated function.

\subsection{Parameterization of the Network}

The parameters characterizing the network comprise all the weights and biases, outlined as follows:

\begin{itemize}
    \item We denote these parameters as \( \phi = \{\textbf{W}^{(h)}, \textbf{b}^{(h)}\}_{h=1}^{H+1}\).
    \item Each layer \(h\) possesses its weight matrix denoted by \(\textbf{W}^{(h)}\) and a bias vector represented by \(\textbf{b}^{(h)}\).
\end{itemize}

Consequently, the network $\mathcal{M}(\textbf{p}; \phi)$ embodies a plethora of parameterized functions, where \(\phi\) demands meticulous selection to ensure the network effectively approximates the target function \(\textbf{f}(\textbf{p})\) at the input \(\textbf{p}\).

\subsection{Gradient Calculation using Back-propagation}

In this section, we present the back-propagation expression \( \frac{\partial \mathcal{L}}{\partial \textbf{W}^{(h)}} \) as follow
(further insights can be found in \cite{Schmidhuber15}):
\begin{equation}\label{bp}
    \frac{\partial \mathcal{L}}{\partial \textbf{W}^{(h)}} = \frac{\partial \mathcal{L}}{\partial \textbf{z}^{(h)}} \cdot \frac{\partial \textbf{z}^{(h)}}{\partial \textbf{W}^{(h)}} = \frac{\partial \mathcal{L}}{\partial \textbf{z}^{(h)}} \otimes \textbf{p}^{(h-1)}.
\end{equation}
where \( \textbf{p}^{(h-1)} \) is the output at the \( h - 1 \) layer and \( \frac{\partial \mathcal{L}}{\partial \textbf{z}^{(h)}} \) is the gradient of the loss with respect to the non-linear operation at layer \( h \). Here, \( [\textbf{p} \otimes \textbf{y}]_{ij} = {\rm p}_i {\rm y}_j \) represents the outer product. Therefore, to evaluate \( \frac{\partial \mathcal{L}}{\partial \textbf{W}^{(h)}} \), both \( \textbf{p}^{(h-1)} \), evaluated during the forward phase, and \( \frac{\partial \mathcal{L}}{\partial \textbf{z}^{(h)}} \), evaluated during back-propagation, are required. Numerical experiments on \eqref{bp} for a specific epoch (iteration) number are presented in Fig.~\ref{Ex1_5}.

\subsection{Training and Testing of MLPs}

In supervised learning, training and testing phases optimize and assess neural networks. Given a dataset \( \mathcal{S} = \{(\textbf{p}_i, \textbf{y}_i) : 1 \leq i \leq n\} \) representing a target function \( \textbf{f} : \textbf{p} \rightarrow \textbf{y} \), the network \( \mathcal{M}(\textbf{p}; \phi, \theta) \) aims to approximate this function, where \( \phi \) are network parameters and \( \theta \) are hyperparameters.
The process includes:
1. \textbf{Training Phase:} Train the network using \( \mathcal{S}_{\text{train}} \) to find optimal parameters \( \phi^* \) by minimizing:
\[
\phi^* = \arg \min_\phi \mathcal{L}_{\text{train}}(\phi) = \frac{1}{n_{\text{train}}} \sum_{i=1}^{n_{\text{train}}} \| \textbf{y}_i - \mathcal{M}(\textbf{p}_i; \phi, \theta) \|_2^2 
\]

2. \textbf{Testing Phase:} Evaluate the optimized network on \( \mathcal{S}_{\text{test}} \) to assess performance on unseen data.
Note that the dataset \( \mathcal{S} \) is split into training and test sets for robust evaluation.


\section{MLPs for Computer Graphics}
\subsection{Algorithm}
In this section, we provide a mathematical foundation for the proposed method that leverages neural networks for surface generation in computer graphics.
\noindent
Let $\mathcal{S}$ represent the 3D domain, and let $\mathbf{p} = \{{\rm{p}}_1, {\rm{p}}_2, \ldots, {\rm{p}}_n\}$ denote the set of data points sampled within this domain. Each data point $\{{\rm{p}}_j\}_{j=1}^n$ has spatial coordinates ${\rm{p}}_j = (x_{1,j}, x_{2,j}, x_{3,j})$.
We categorize the data points into three sets:
\begin{align*}
    \mathbf{p}_\text{surface} & : \text{Data points defining the surface boundary (Fig.~\ref{algorithm}, step 1)} \\
        \mathbf{p}_\text{interior} & : \text{Data points located within the 3D object (Fig.~\ref{algorithm}, step 2)} \\
    \mathbf{p}_\text{exterior} & : \text{Data points outside the object (only considered for Example 4)}
\end{align*}
\noindent
We define $n_i$, $n_s$, and $n_e$ as the number of points inside, on the surface, and outside of the domain $\mathcal{S}$. We also consider $n$ as the total number of data points, where $n = n_i + n_s + n_e$.

To train the neural network, we label each data point $\mathbf{p}_j$ as follows:
\begin{align}\label{labels}
    \text{Label}({\rm{p}}_j) &= 
    \begin{cases}
        1 & : {\rm{p}}_j \in \mathbf{p}_\text{interior} \\
        0 & : {\rm{p}}_j \in \mathbf{p}_\text{surface} \\
        -1 & : {\rm{p}}_j \in \mathbf{p}_\text{exterior}
    \end{cases}
\end{align}
\noindent
The proposed neural network architecture consists of a combination of feedforward and highway networks, with the power-enhanced version. It is designed to learn the distinct features associated with each category of data points.
\noindent
The loss function, chosen as the mean squared error (MSE), quantifies the difference between predicted labels and actual labels. It is expressed as:
\begin{equation}
    \text{MSE} = \frac{1}{n} \sum_{i=1}^{n} \left(\text{Label}({\rm{p}}_i) - \text{Prediction}({\rm{p}}_i)\right)^2
\end{equation}
where $n$ is the number of data points (samples), $\text{Label}({\rm{p}}_i)$ represents the ground truth label, and $\text{Prediction}(\rm{p}_i)$ is the network's predicted label for the input data point ${\rm{p}}_i$.
\noindent
The network's parameters, including weights and biases, are optimized using an iterative optimization algorithm, L-BFGS-B (Limited-memory Broyden-Fletcher-Goldfarb-Shanno with Box constraints), to minimize the MSE (Fig.~\ref{algorithm}, step 3). This process equips the network with the ability to accurately predict labels for new data points and, subsequently, to generate surfaces for 3D objects in computer graphics.

\subsection{Testing and Surface Generation Algorithm}

\begin{figure}[!h]
\centering%
\includegraphics[width=5.95in]{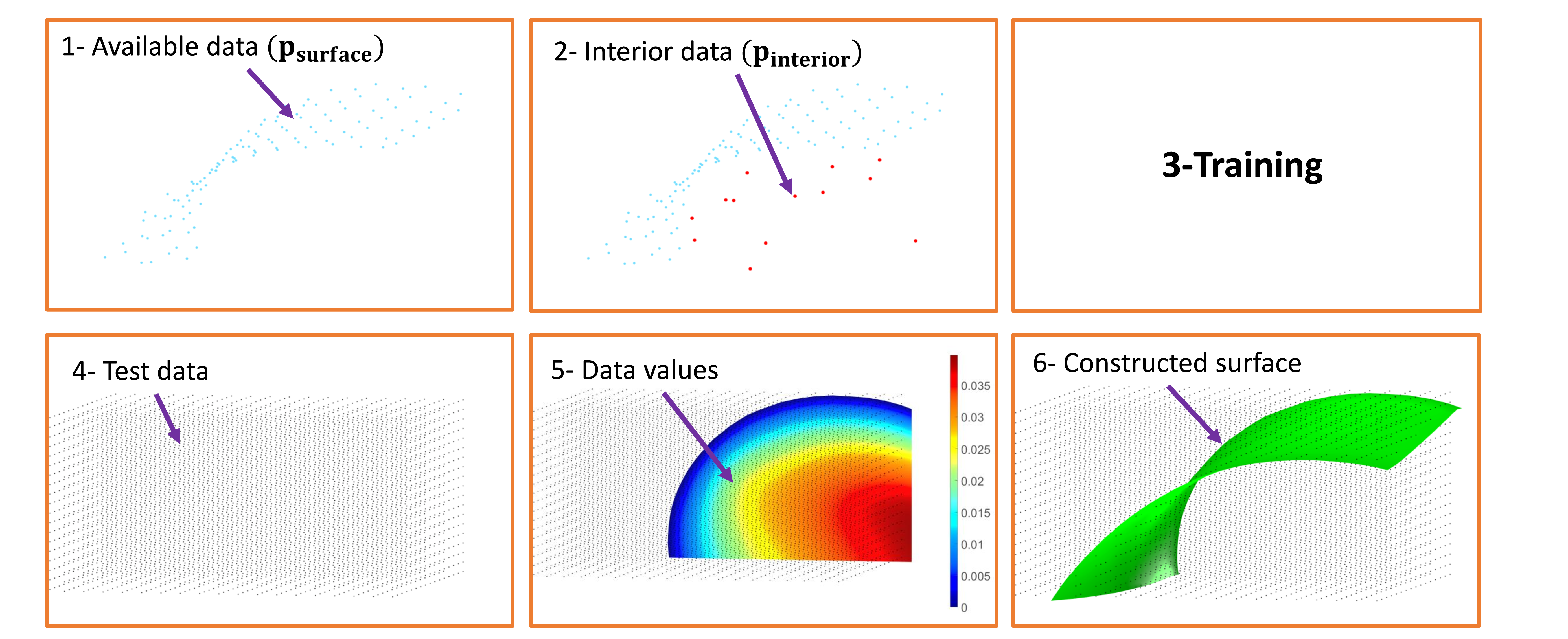}
\caption{Algorithm to construct a surface from available data. } \label{algorithm}
\end{figure}

After training the neural network, the next step is to apply the model to generate surfaces for 3D objects. This involves testing the network's ability to predict labels for a set of testing data points. In our case, we employ a grid-based approach to define the testing data points over the 3D domain $\mathcal{S}$.
Let $\mathcal{S}_{test} = \{\mathbf{t}_1, \mathbf{t}_2, \ldots, \mathbf{t}_m\}$ represent the set of testing data points, where each point $\mathbf{t}_j$ corresponds to a meshgrid point over the 3D domain (Fig.~\ref{algorithm}, step 4). We use these points to evaluate the network's performance in surface generation.

To visualize the surfaces, we utilize the Matlab {\tt{``isosurface''}} command. This command allows us to create contours over the 3D domain (Fig.~\ref{algorithm}, step 6) based on the predicted values from the neural network (Fig.~\ref{algorithm}, step 5). Specifically, we set the isosurface threshold value to ``0'' effectively marking the boundary of the generated surface.
In this way, we obtain a 3D contour that visually represents the generated surface for the 3D object. By selecting a threshold of ``0'' we ensure that the contour encircles the surface points, which were labeled as ``0'' during the training process \eqref{labels}. These steps are shown in Fig.~\ref{algorithm}.

\section{Advanced Deep Neural Network Architectures}
\subsection{Highway Networks}
\begin{figure}[!h]
\centering
\includegraphics[width=6.3in]{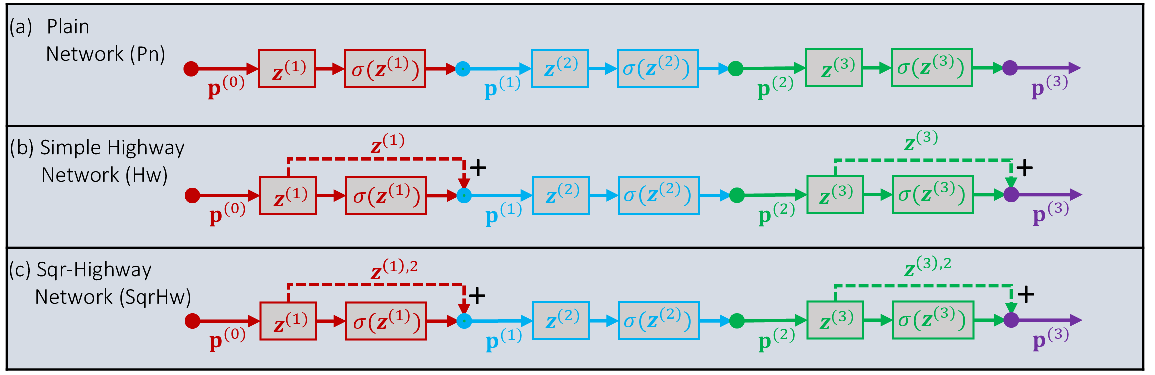}
\caption{Three neural network architectures: (a) a basic neural network (Plain net), (b)  the proposed simple highway network (Hw) (c) Square-Highway network (SqrHw).} 
\label{ExNet}
\end{figure}

This section introduces Highway Networks, a type of deep neural network designed to improve information propagation across multiple layers \cite{Srivastava15a,Srivastava15}. Highway Networks use gating mechanisms to control information flow, allowing selective processing and preservation of input data. A conventional neural network can be represented as (Fig.~\ref{ExNet}(a)):
\begin{equation}
\textbf{p}^{(h)} = f^{(h)}(\textbf{p}^{(h-1)}, \textbf{W}_{f}^{(h)})
\end{equation}
where \( \textbf{p} \) is the input to the layer and \( f^{(h)}(\textbf{p}^{(h-1)}, \textbf{W}_{f}^{(h)}) \) represents the operations at layer \( h-1 \).
Highway Networks add two gating mechanisms:
\begin{itemize}
    \item Nonlinear transformations, controlled by the transform gate \( T \),
    \item Activation transfer from the previous layer, managed by the carry gate \( C \).
\end{itemize}
This is represented by:
\begin{equation}\label{hn}
\textbf{p}^{(h)} = f^{(h)}(\textbf{p}^{(h-1)}, \textbf{W}_{f}^{(h)})\cdot T^{(h)}(\textbf{p}^{(h-1)}, \textbf{W}_{T}^{(h)}) + \textbf{p}^{(h-1)}\cdot C^{(h)}(\textbf{p}^{(h-1)}, \textbf{W}_{C}^{(h)})
\end{equation}
where \( f^{(h)}(\textbf{p}^{(h-1)}) \) includes the linear and nonlinear functions within the network. This allows Highway Networks to learn both feedforward and shortcut connections, aiding the training of very deep networks and preserving crucial information. Highway Networks have shown promising results in tasks like speech recognition, natural language processing, and image classification \cite{Srivastava15a,Srivastava15}. However, this architecture require additional training for the carry and transform gates, which can be very costly. Another challenge in highway networks is dimensionality adjustment. 
Equation \eqref{hn} requires the dimensions of $\mathbf{p}^{(h)}$, $T^{(h)}(\mathbf{p}^{(h-1)}, \mathbf{W}_{T}^{(h)})$, and $C^{(h)}(\mathbf{p}^{(h-1)}, \mathbf{W}_{C}^{(h)})$ to be consistent. This can be managed by:
\begin{itemize}
    \item Sub-sampling or zero-padding $\mathbf{p}^{(h)}$ to get $\hat{\mathbf{p}}^{(h)}$,
    \item Using a plain layer to adjust the dimensionality before stacking highway layers.
\end{itemize}
The following networks address and resolve the issues of extra training and dimensionality.

\subsection{Residual Network}

Residual Networks present a simplified version of  Highway Networks \cite{Greff17}, where the transformation is redefined as the sum of the input and a residual. Commonly referred to as ResNets \cite{He15, He16}, these networks have become prominent in neural network architectures.These networks are characterized by their residual modules, denoted as \( f^{(h)} \), and skip connections that circumvent these modules, aiding in the creation of deep networks. This allows for the establishment of residual blocks, which consist of collections of layers within the network.
 Hence, the output \(\textbf{p}^{(h)}\) for the \(h\)-th layer is defined as:
\begin{equation}
\textbf{p}^h = f^{(h)}(\textbf{p}^{(h-1)}, \textbf{W}_{f}^{(h)}) + \textbf{p}^{(h-1)}. \label{eq:residual}
\end{equation}
In this context, \( f^{(h)}(\textbf{p}^{(h-1)}, \textbf{W}_{f}^{(h)}) \) encapsulates a sequence of computations, comprising linear transformations \eqref{eq1} and element-wise activation functions \eqref{eq2} at layer \(h-1\), for residual networks where \( 1\leq h \leq H \). A comparison between the residual network given by \eqref{eq:residual} and the highway network represented by \eqref{hn} reveals that both the transfer $(T)$ and carry $(C)$ gates are set to 1 \cite{Greff17}, i.e.
\begin{equation}
T^{(h)}(\textbf{p}^{(h-1)}, \textbf{W}_{T}^{(h)}) = 1
\end{equation}
and 
\begin{equation}
\textbf{p}^{(h-1)}\cdot C^{(h)}(\textbf{p}^{(h-1)}, \textbf{W}_{C}^{(h)}) = \textbf{p}^{(h-1)}
\end{equation}
implying no additional updates are necessary.

\subsection{Proposed Simple Highway Network (Hw)}
Here, we introduce a simplified version of the highway network, represented by Expression \eqref{hn} as shown in Fig.~\ref{ExNet}(b):
\begin{equation}
\textbf{p}^{(h)} = f^{(h)}(\textbf{p}^{(h-1)}, \textbf{W}_{f}^{(h)})+ \textbf{Z}^{(h)}  \label{eq:simplehw}
\end{equation}
In this equation, $\textbf{Z}^{(h-1)} = \textbf{W}_{f}^{(h)} {\textbf{p}}^{(h-1)} + \textbf{b}^{(h)}$.
To compare this simplified Highway network \eqref{eq:simplehw} with the original one in \eqref{hn}, we observe that in the simplified version, we set 
\begin{equation}
T^{(h)}(\textbf{p}^{(h-1)}, \textbf{W}_{T}^{(h)}) = 1
\end{equation}
and 
\begin{equation}
\textbf{p}^{(h-1)}\cdot C^{(h)}(\textbf{p}^{(h-1)}, \textbf{W}_{C}^{(h)}) = \textbf{W}^{(h)}_{f} {\textbf{p}}^{(h-1)} + \textbf{b}^{(h)}
\end{equation}
This implies that the weight updates for both $\mathbf{W}^{(h)}_{f}$ and $\mathbf{W}^{(h)}_{c}$ in the simplified Highway network are identical, eliminating the necessity for additional training in contrast with the original highway network. Additionally, in this case, no matrix size alignment is required since the weight of the plain network is considered.

\subsection{Proposed Square-Highway Network (SqrHw)}
Here, we propose a simple highway network, where \eqref{hn} can be represented as (Fig.~\ref{ExNet}(c)):
\begin{equation}
\textbf{p}^{(h)} = f^{(h)}(\textbf{p}^{(h-1)}, \textbf{W}_{f}^{(h)})+ \textbf{Z}^{{(h)}} \odot \textbf{Z}^{{(h)}}  \label{eq:sqrhw}
\end{equation}
where $\odot$ denotes element-wise multiplication. In this equation, $\textbf{Z}^{(h)} = \textbf{W}_{f}^{(h)} {\textbf{p}}^{(h-1)} + \textbf{b}^{(h)}$.
To compare the proposed SqrHw  network represented by \eqref{eq:sqrhw} with the original one in \eqref{hn}, we note that in the simplified version, we set
\begin{equation}
T^{(h)}(\textbf{p}^{(h-1)}, \textbf{W}_{T}^{(h)}) = 1
\end{equation} 
and the term
\begin{equation}\label{sqrhwcarry}
\textbf{p}^{(h-1)}\cdot C^{(h)}(\textbf{p}^{(h-1)}, \textbf{W}_{C}^{(h)}) = (\textbf{W}^{(h)}_{f} {\textbf{p}}^{(h-1)} + \textbf{b}^{(h)}) \odot ( \textbf{W}^{(h)}_{f} {\textbf{p}}^{(h-1)} + \textbf{b}^{(h)})
\end{equation}
Here, the operation $\odot$ represents element-wise multiplication, implying that the carry gate applies the same transformation to the input $\textbf{p}^{(h-1)}$ as the original Highway network, but with additional element-wise modification.
Equation \eqref{sqrhwcarry} suggests that the weight updates are identical $(\textbf{W}^{(h)}_{f} = \textbf{W}^{(h)}_{c})$ to those of a plain neural network, thereby removing the necessity for additional training or dimensionality alignment.

\subsection{Development of Residual and Highway Schemes in Neural Networks}
Works like the Highway Network and ResNet initially lacked rigorous mathematical proofs and relied heavily on empirical validation through extensive numerical experiments once they were introduced. Despite this, they have become widely adopted due to their empirical success, underscoring the importance of practical utility alongside mathematical rigor.
Or well-cited works such as References \cite{LuLu20,Wang21} introduced completely different residual schemes to the network. For instance, the authors in \cite{LuLu20} introduce the residual connection from low-fidelity output to high-fidelity output, which will be implemented once in every epoch or iteration (\cite{LuLu20}, Figure 1D). In reference \cite{Wang21}, an elementwise multiplication operation is applied between the residual terms and the plain neural network's output (\cite{Wang21}, Equation 2.36). Although both of these methods diverge from the original ResNet and highway architectures \cite{Srivastava15, He15}, they have proven to be successful approaches.


\section{Numerical Analysis}

In the context of this study, we make use of the notations \(n\), \(n_l\), and \(n_n\) to describe the count of data points, layers, and neurons in each layer, respectively. Additionally, \(n_i\), \(n_s\),  \(n_e\), and \(n_t\) are employed to denote the number of interior, surface, exterior, and validation data points. In this section, an examination of three distinct approaches is undertaken:

\begin{enumerate}
\item \texttt{Plain Network (Pn)}: A standard neural network lacking additional modifications or residual connections (refer to Fig.~\ref{ExNet}(a)).

\item \texttt{Simple Highway (Hw)}: This neural network architecture includes affine transformations that are added to the output of every other layer (illustrated in Fig.~\ref{ExNet}(b)). The rationale for adding to every other layer rather than every layer is for efficiency purposes.

\item \texttt{Squared Highway (SqrHw)}: An innovative variation of the Highway network architecture, introducing squared affine transformation every other layer (depicted in Fig.~\ref{ExNet}(c)). The reason for adding to every other layer instead of every layer is for efficiency.
\end{enumerate}
\noindent
In all following examples, we exclusively considered interior points $\mathbf{P}_\text{interior}$ in \eqref{labels}. In the last example, the Stanford Bunny, we incorporated both interior  $\mathbf{P}_\text{interior}$ and exterior points $\mathbf{P}_\text{exterior}$ for the surface reconstruction.

\begin{example}\rm{

In our first numerical example, we examine the construction of a simple and smooth surface, a sphere. The sphere has a radius of 1 and is centered at the origin (0, 0, 0), as depicted in Fig.~\ref{Ex1_sphere_pts}. We employ $n_s=200$ data points on the sphere's surface from  $\mathbf{P}_\text{interior}$, along with $n_i=20$ interior points from $\mathbf{P}_\text{surface}$, without any exterior points from $\mathbf{P}_\text{exterior}$. The points are randomly selected. Fig.~\ref{Ex0_2} displays the results of the surface construction over these points. The green surfaces represent the outputs of the neural network, while the red meshed spheres are the exact surface which provided for reference to enhance the visualization of the results.

%

\begin{figure}[!h]
\centering%
\includegraphics[width=3.35in]{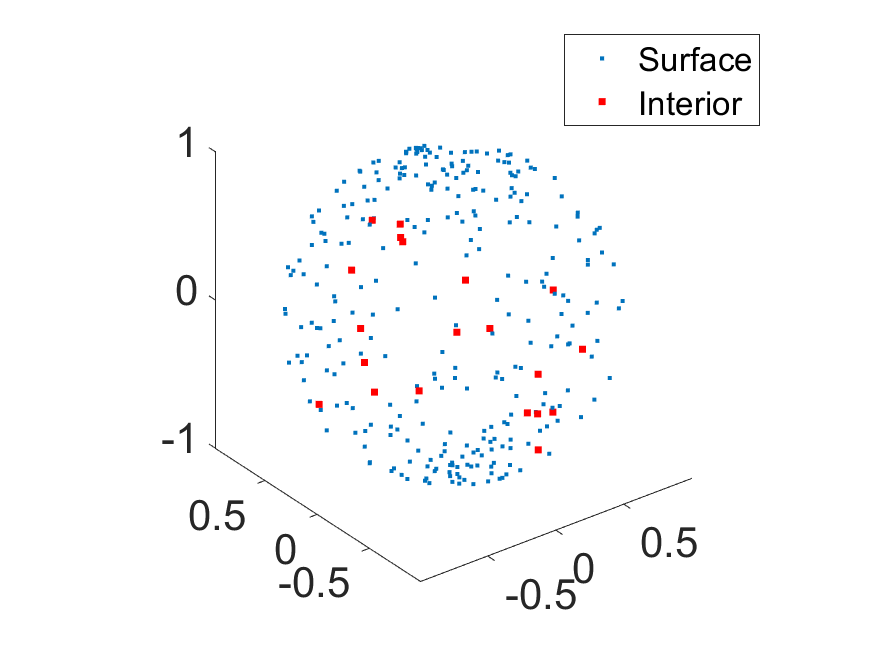}
\caption{The profile of the point distribution for the sphere. } \label{Ex1_sphere_pts}
\end{figure}

\begin{figure}[!h]
\centering%
\subfigure[]{ \label{Ex1_pn_iter10}
\includegraphics[width=1.45in]{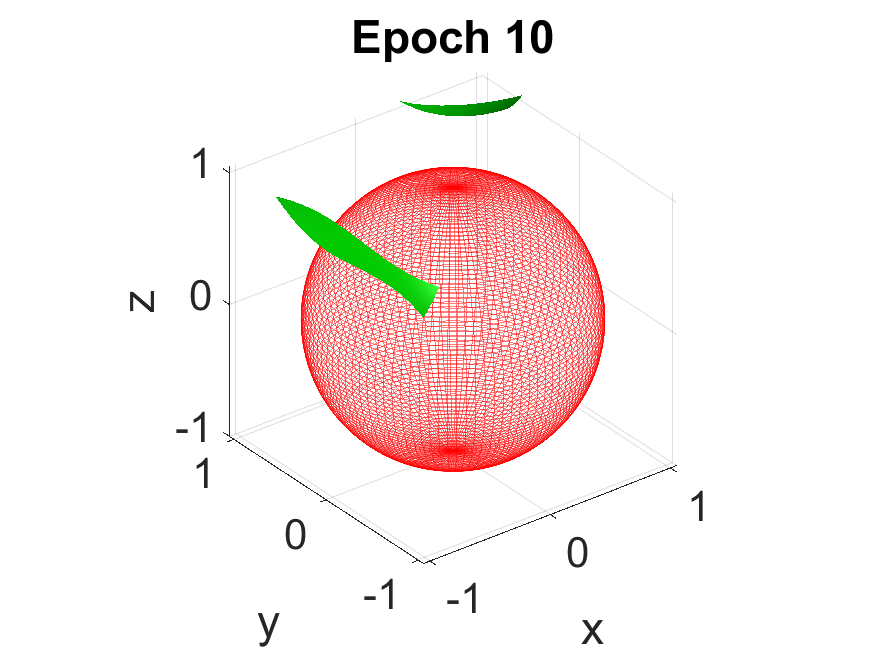}}
\subfigure[]{ \label{Ex1_pn_iter50}
\includegraphics[width=1.45in]{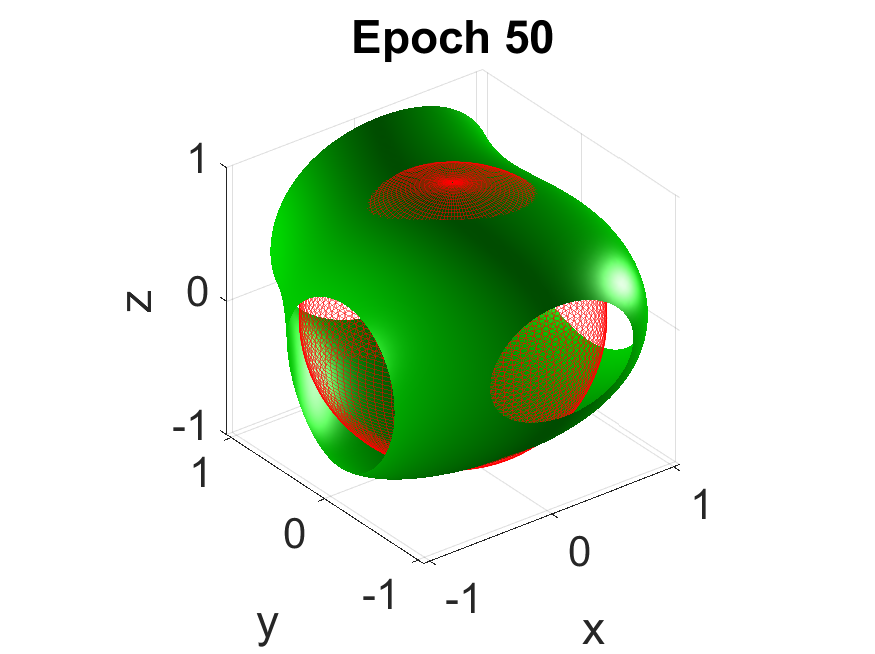}}
\subfigure[]{ \label{Ex1_pn_iter100}
\includegraphics[width=1.45in]{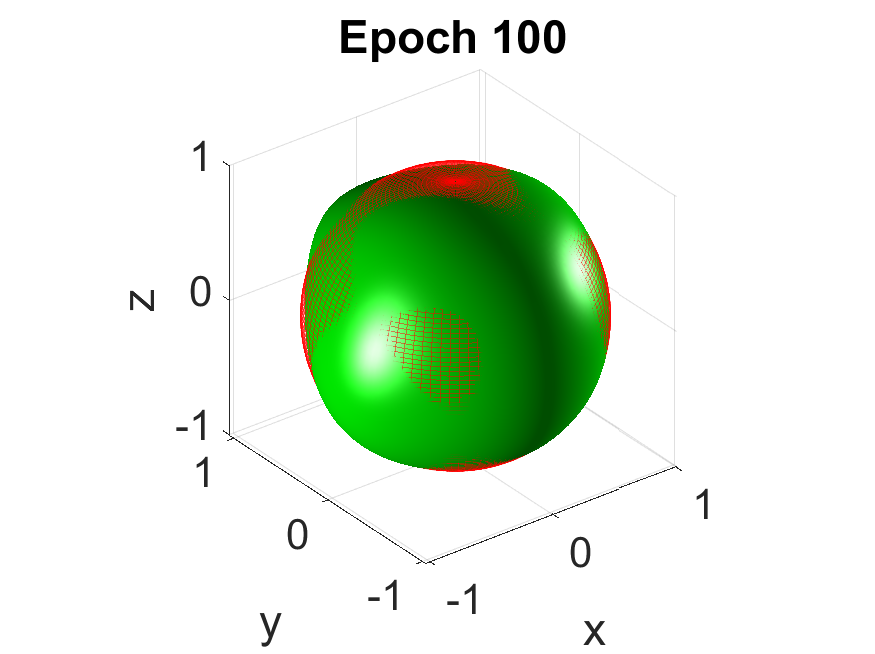}}
\subfigure[]{ \label{Ex1_pn_iter400}
\includegraphics[width=1.45in]{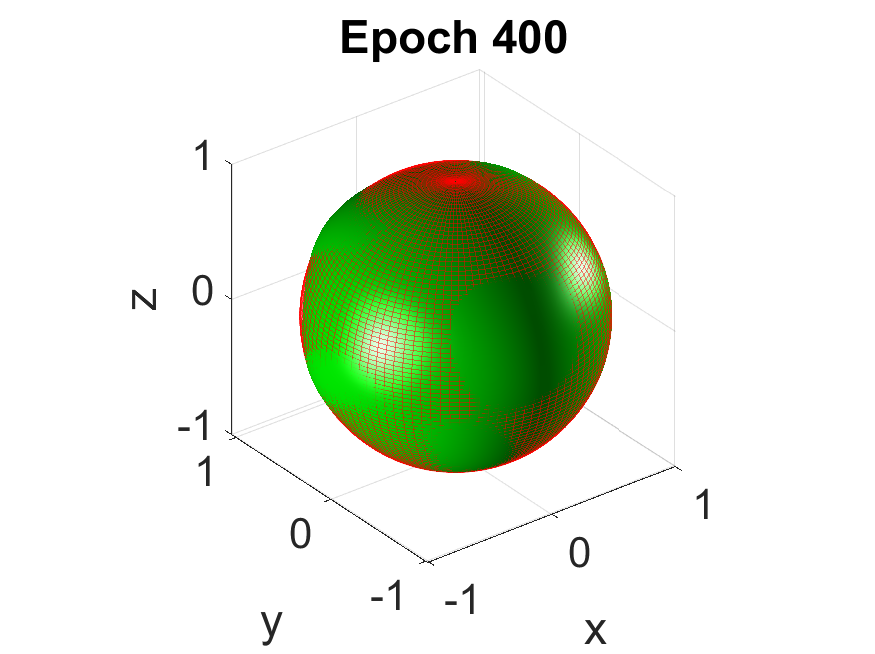}}
\subfigure[]{ \label{Ex1_res_iter10}
\includegraphics[width=1.45in]{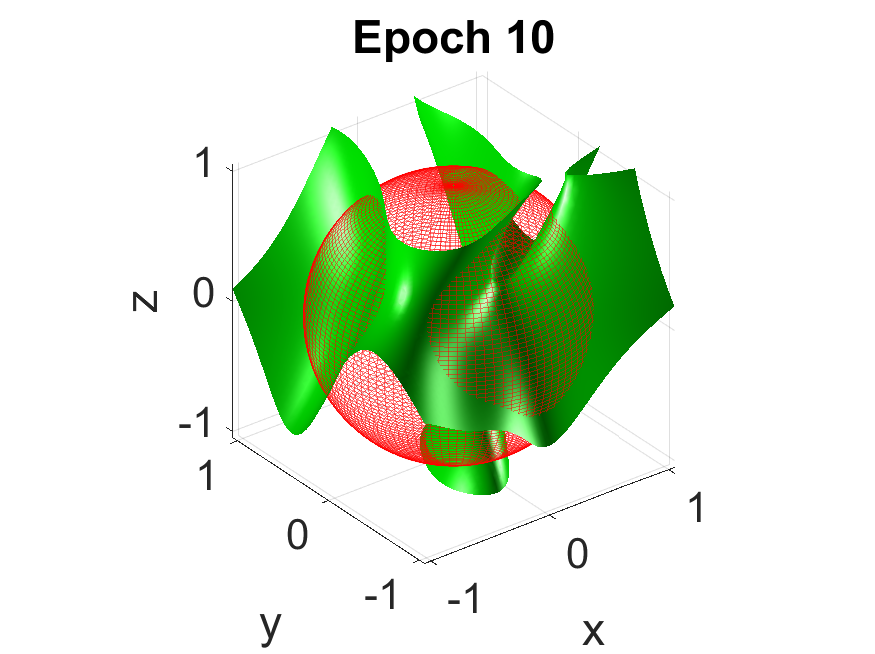}}
\subfigure[]{ \label{Ex1_res_iter50}
\includegraphics[width=1.45in]{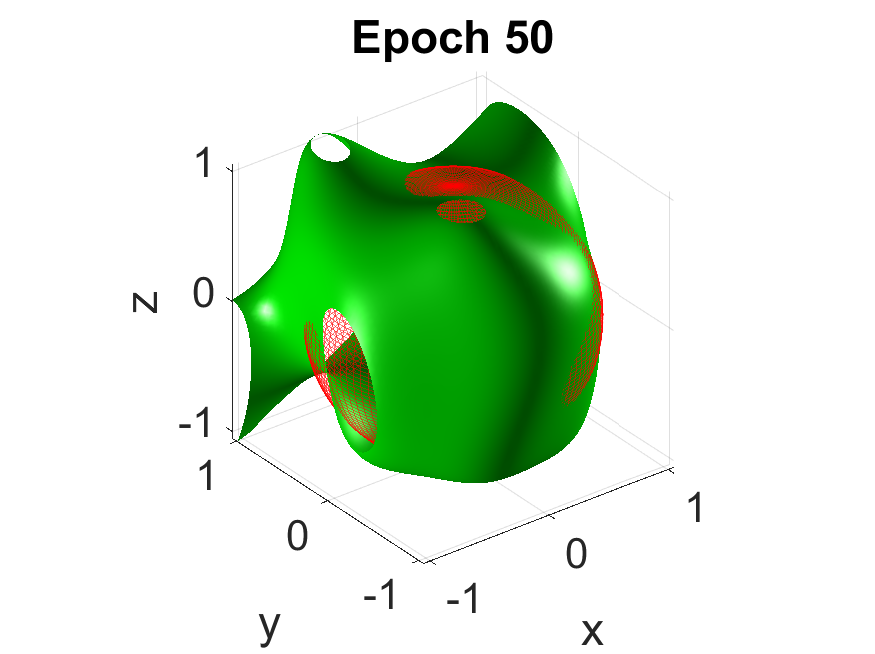}}
\subfigure[]{ \label{Ex1_res_iter100}
\includegraphics[width=1.45in]{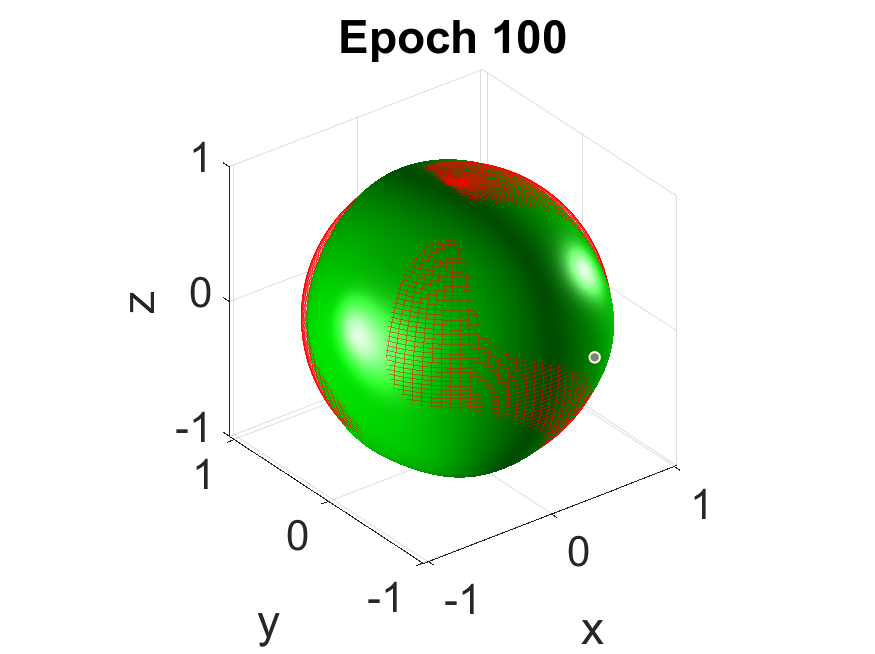}}
\subfigure[]{ \label{Ex1_res_iter400}
\includegraphics[width=1.45in]{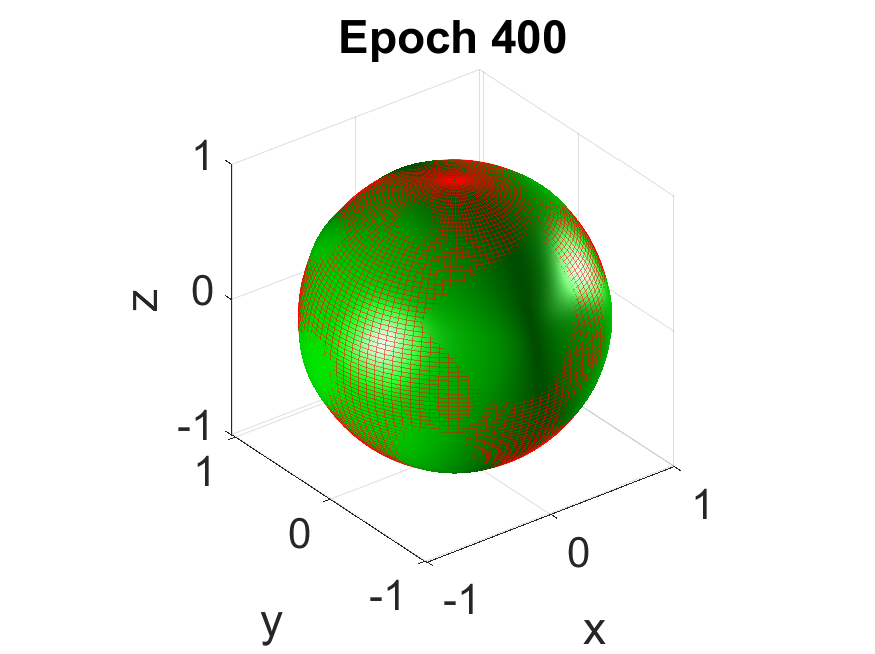}}
\subfigure[]{ \label{Ex1_sqr_iter10}
\includegraphics[width=1.45in]{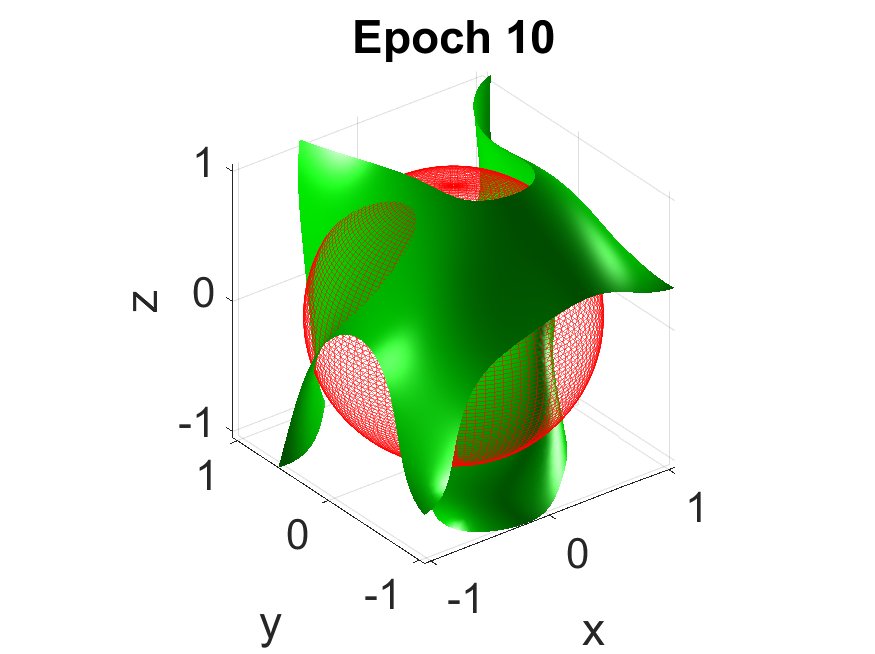}}
\subfigure[]{ \label{Ex1_sqr_iter50}
\includegraphics[width=1.45in]{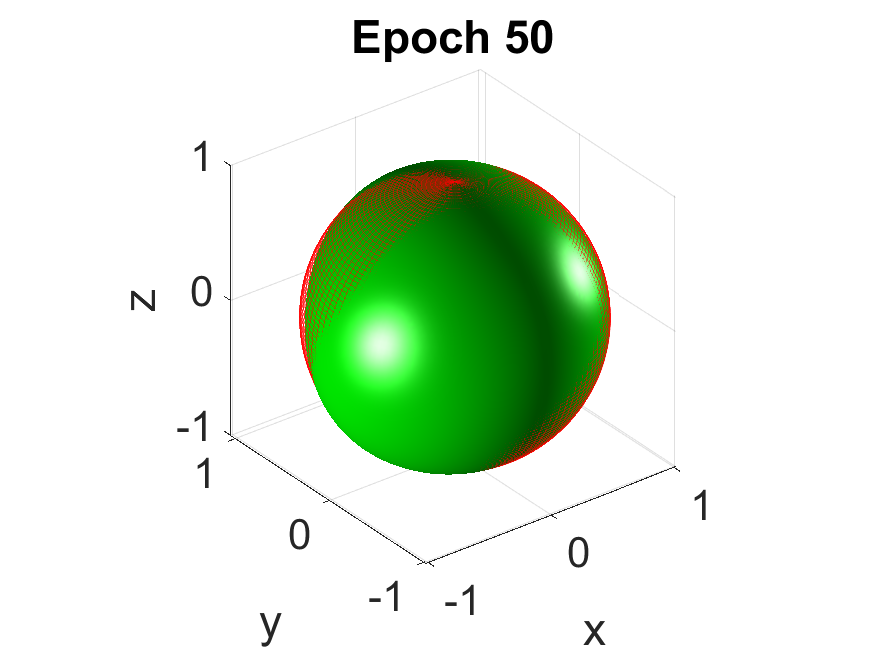}}
\subfigure[]{ \label{Ex1_sqr_iter100}
\includegraphics[width=1.45in]{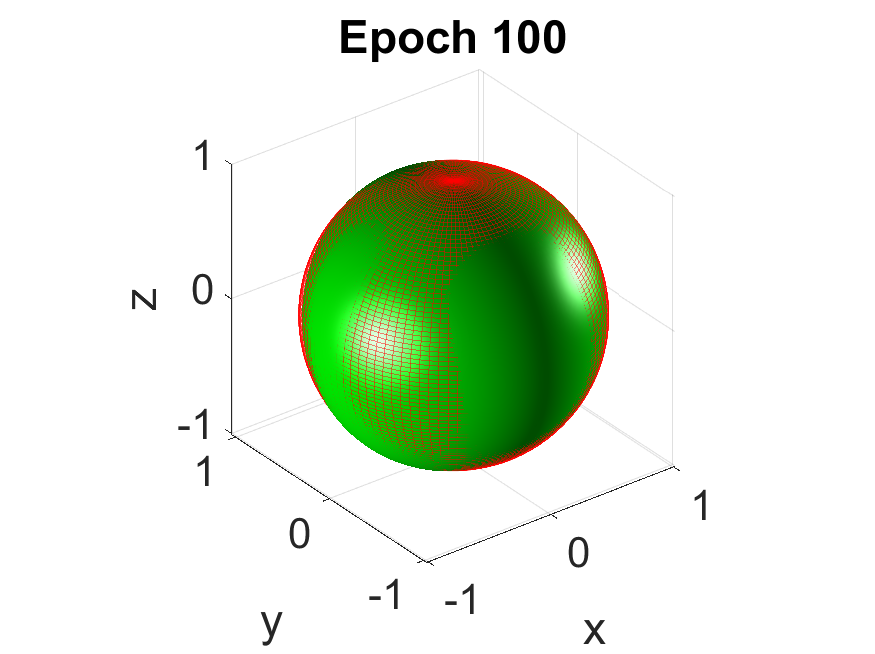}}
\subfigure[]{ \label{Ex1_sqr_iter400}
\includegraphics[width=1.45in]{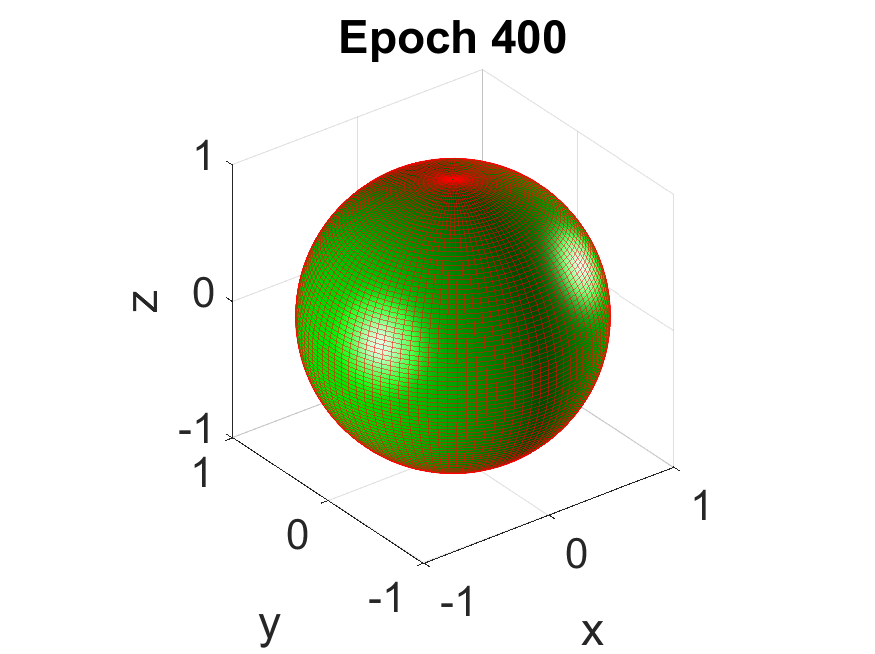}}
\caption{The profile of the simulated sphere. The top panel: plain network,  middle panel:  Hw, bottom panel: SqrHw. } \label{Ex0_2}
\end{figure}

\noindent
The first row of Figure \ref{Ex0_2} illustrates the results using the plain network for different epochs, ranging from 10 to 400. The second row showcases the results for the same epochs, employing the simple Highway approach. Finally, the last row demonstrates the results for the proposed square highway.

Several observations can be made from this example:

\begin{itemize}
  \item The proposed SqrHw demonstrates superior performance at specific iterations, especially when compared to the plain network. In other words, it converges faster when evaluating the constructed surface at specific epochs compared to the other networks.
  \item The Hw exhibits better performance than the plain network.
  \item At the final epoch, we observe that the SqrHw is very close to the reference sphere (red meshed sphere), while the other two networks have not yet achieved an accurate representation.
\end{itemize}

\begin{figure}[!h]
\centering%
\includegraphics[width=5.95in]{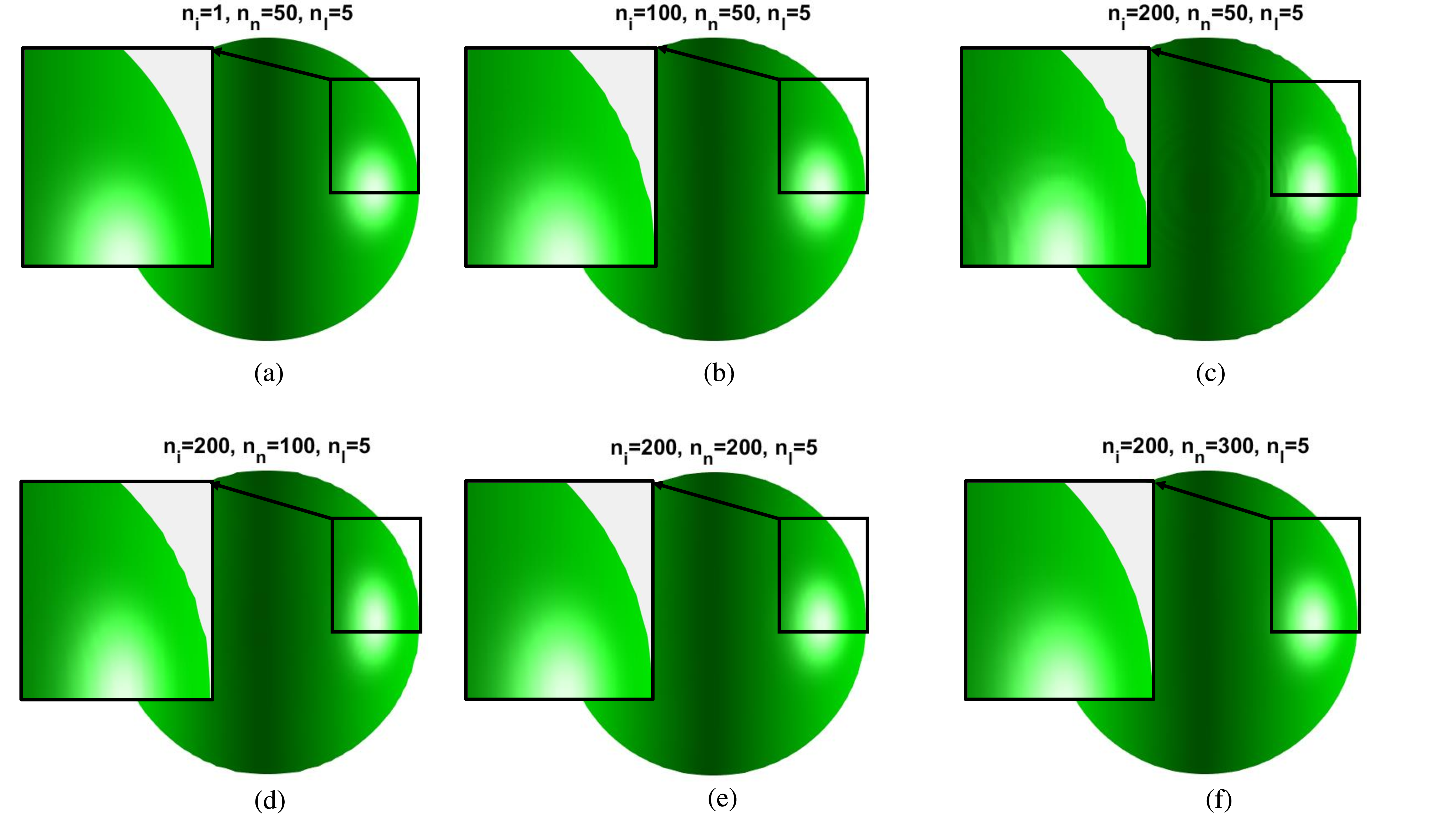}
\caption{The profile of the simulated sphere for $n=200$ using SqrHw for various number of interior points (top panel) and neurons (bottom panel.) } \label{Ex0_3}
\end{figure}

In conclusion, the proposed method outperforms the other approaches. 
Next, we investigate the impact of the number of interior points ($n_i$) and the number of neurons in each hidden layer ($n_n$) in our work. For this analysis, we exclusively consider the proposed SqrHw. Figure \ref{Ex0_3} displays the results of constructing the sphere in 2D for various numbers of interior points (top row) and various numbers of hidden layers (bottom row).
In the top row, starting with just one interior point (plot \ref{Ex0_3}(a)), we observe that the proposed method performs well even with this minimal number of interior points. As we increase the number of interior points to 100 (plot \ref{Ex0_3}(b)), we notice that the sphere's edge becomes less smooth. After selecting 200 interior points (plot \ref{Ex0_3}(c)), in the top-right corner, we observe that the edges become even less smooth, with some wave-like artifacts appearing on the sphere's surface.
We maintain the settings at 200 interior points and 5 hidden layers and vary the number of neurons ($n_n$) to examine this parameter's impact on the constructed surface, as shown in the bottom row. As we increase the number of neurons from 50 (plot \ref{Ex0_3}(c)) to 100 (plot \ref{Ex0_3}(d)), we observe that the sphere's surface becomes smoother. Further increasing the number of neurons to 200 (plot \ref{Ex0_3}(e)) and finally to 300 (plot \ref{Ex0_3}(f)) results in a notably smooth and improved surface.

A comparison between the top-right (plot \ref{Ex0_3}(c)) and bottom-right corners (plot \ref{Ex0_3}(f)) demonstrates that increasing the number of neurons is beneficial for constructing a smoother and more accurate surface.
However the selection of the interior points should be done by care. 

}
\end{example}

\begin{example}\rm{

In this example, we delve into a more intricate surface: the hand. Additionally, we explore the details concerning weight updating and backpropagation in this example. Fig.~\ref{Ex_hand_pts} illustrates a node distribution where points on the hand's surface are depicted in blue, and the smaller interior points are shown in red.


\begin{figure}[!h]
\centering%
\includegraphics[width=3.35in]{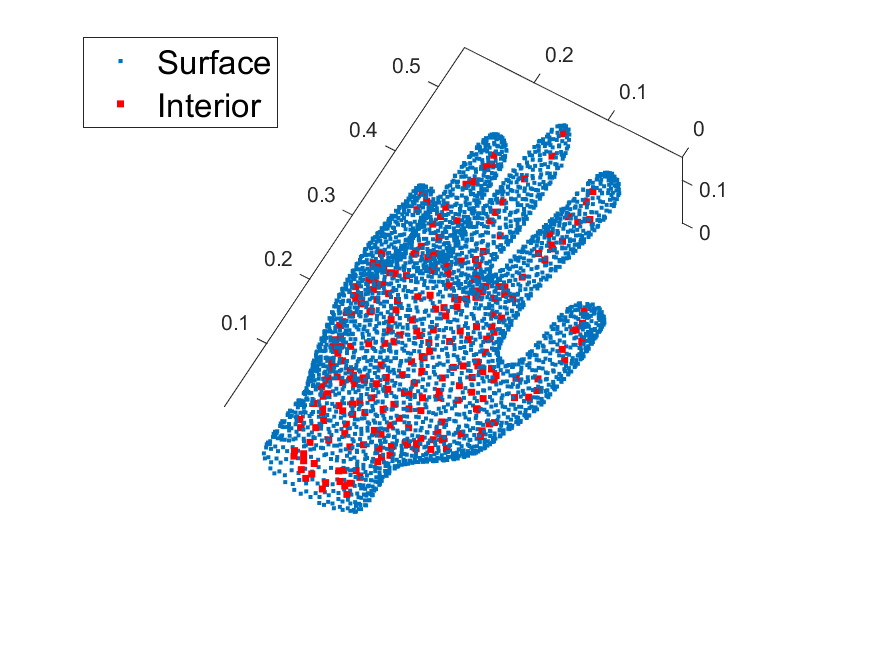}
\caption{The profile of the  point distribution for the hand.  } \label{Ex_hand_pts}
\end{figure}

Moving on to the numerical results, Fig.~\ref{Ex1_2} showcases the outcomes of three different network structures for 3000 points on the surface, with 500 interior points. The structure of the network comprises five hidden layers, each containing 50 neurons. The top panel corresponds to the plain network, the middle panel displays the results of the Hw algorithm, and the bottom panel exhibits the results of our proposed SqrHw network. Within each panel, columns represent different Epoch numbers, specifically 50, 5000, and 10000 Epochs, facilitating easy comparison of convergence across the three network architectures. The final column, Column 4, displays the results at the final epoch where convergence occurred. Several observations can be made from this figure:

\begin{itemize}
  \item At a small number of Epochs, such as 50 as depicted in the first column, the construction begins as a simple plate for all three network structures.
  \item Subsequently, better convergence can be observed with an increasing number of Epochs, particularly for Hw and SqrHw, in contrast to the plain neural network. For example, a comparison between plots in Fig.~\ref{Ex1_2}(c,g,k) reveals the emergence of a recognizable ``hand'' shape for Hw and SqrHw, while the Plain Neural Network produces non-meaningful results.
  \item The final results, shown in the last column, reveal a near-perfect hand shape for SqrHw, as seen in Fig.~\ref{Ex1_2}(l). In contrast, the plain Neural Network's results, as illustrated in Fig.~\ref{Ex1_2}(d), are highly polluted, and Hw's outcomes, Fig.~\ref{Ex1_2}(h), feature some extraneous surface artifacts around the hand.
  \item Notably, the plain neural network requires a substantially smaller number of Epochs (13250) for convergence over training data, while Hw exhibits the longest convergence duration at 21100 Epochs. Conversely, SqrHw achieves its final results at Epoch 18000, positioning it between the plain network and highway in terms of the number of required Epochs for convergence over training data.
\end{itemize}

\begin{figure}[!h]
\centering%
\subfigure[]{ \label{Ex_hand_pn_iter50}
\includegraphics[width=1.45in]{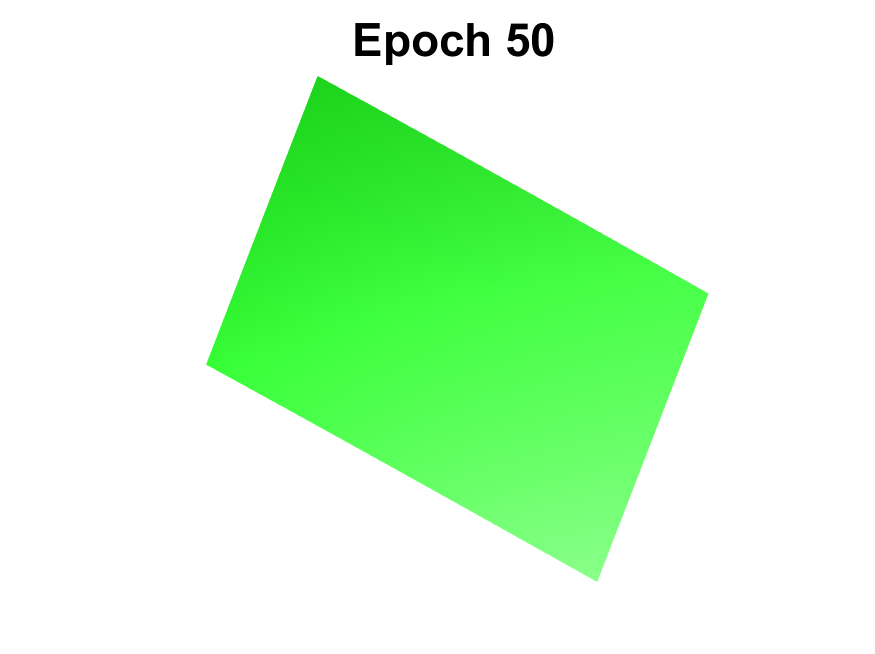}}
\subfigure[]{ \label{Ex_hand_pn_iter5000}
\includegraphics[width=1.45in]{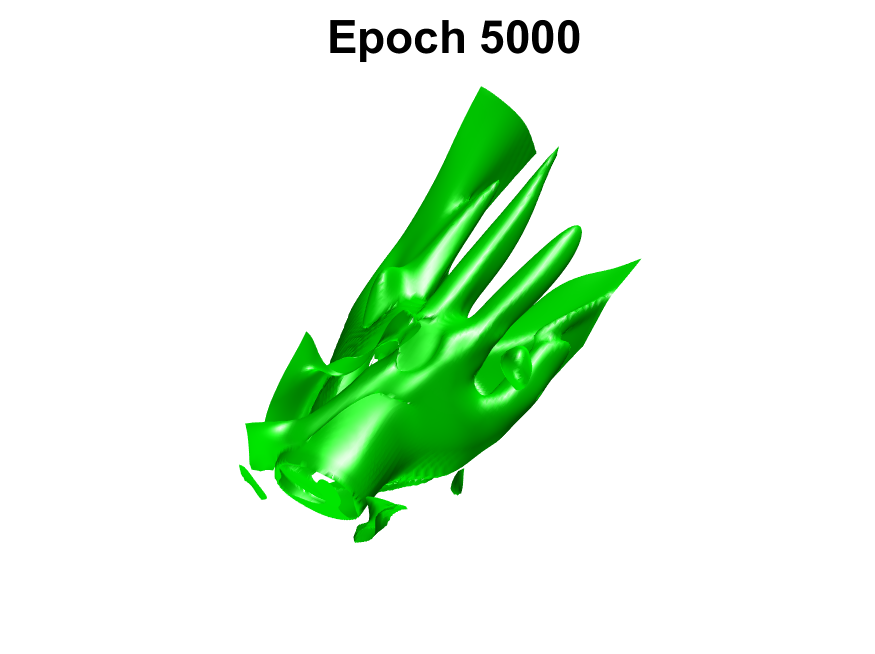}}
\subfigure[]{ \label{Ex_hand_pn_iter10000}
\includegraphics[width=1.45in]{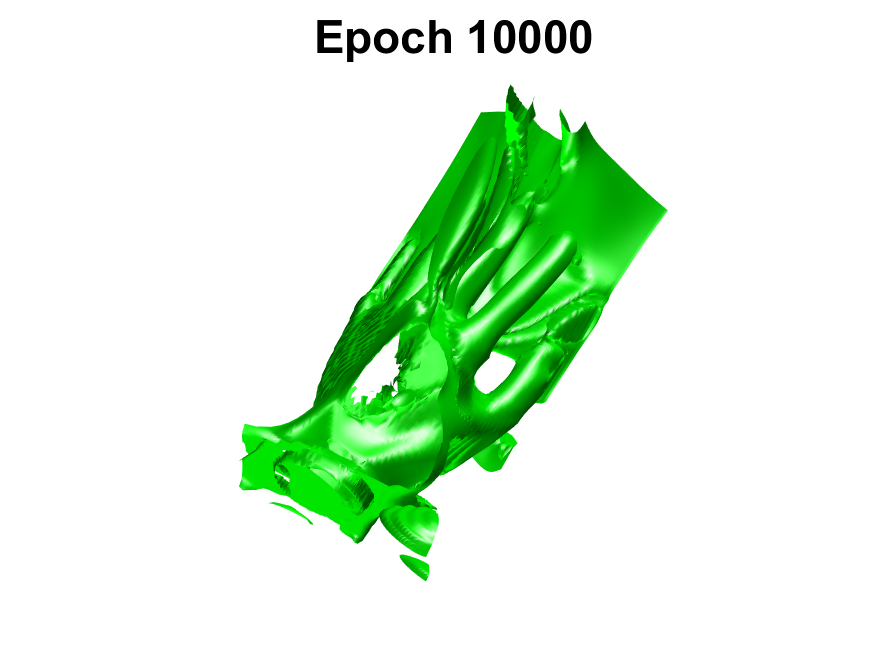}}
\subfigure[]{ \label{Ex_hand_pn_iter13250}
\includegraphics[width=1.45in]{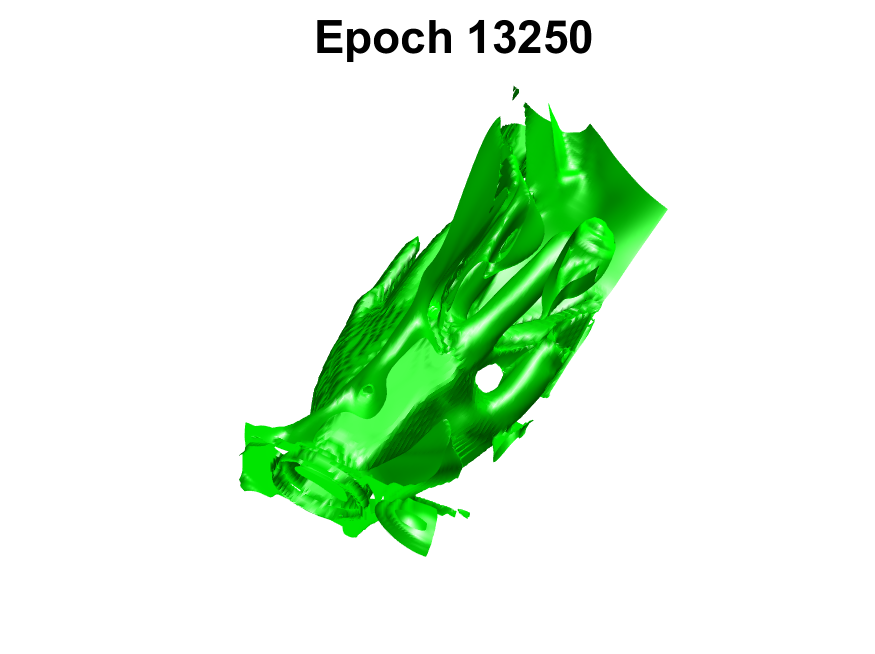}}
\subfigure[]{ \label{Ex_hand_resnet_iter50}
\includegraphics[width=1.45in]{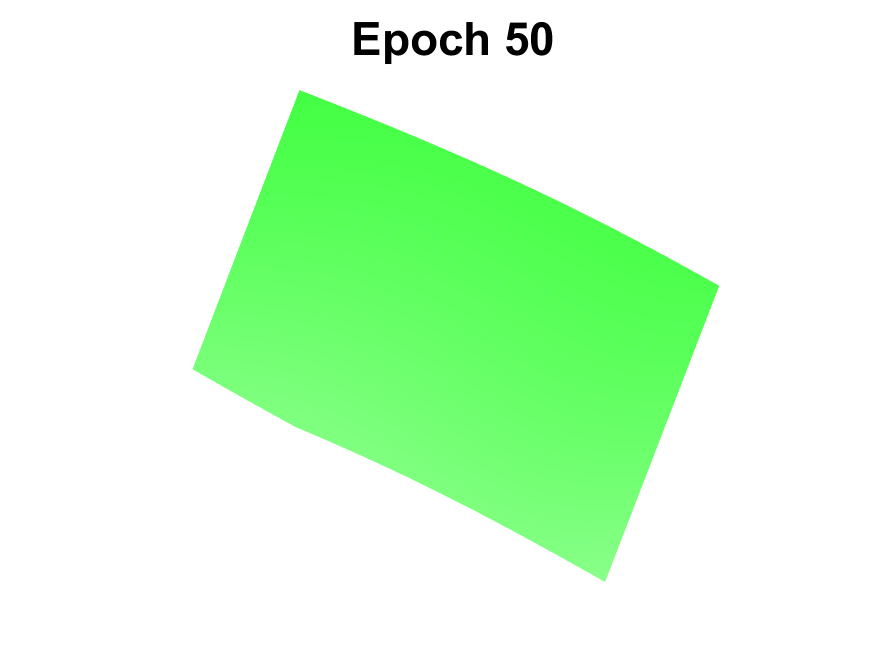}}
\subfigure[]{ \label{Ex_hand_resnet_iter5000}
\includegraphics[width=1.45in]{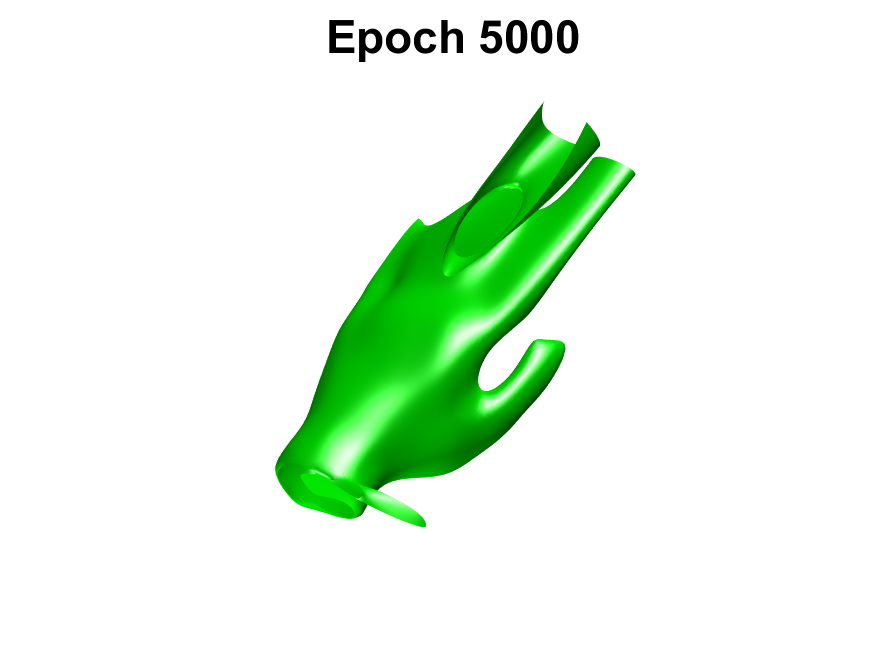}}
\subfigure[]{ \label{Ex_hand_resnet_iter10000}
\includegraphics[width=1.45in]{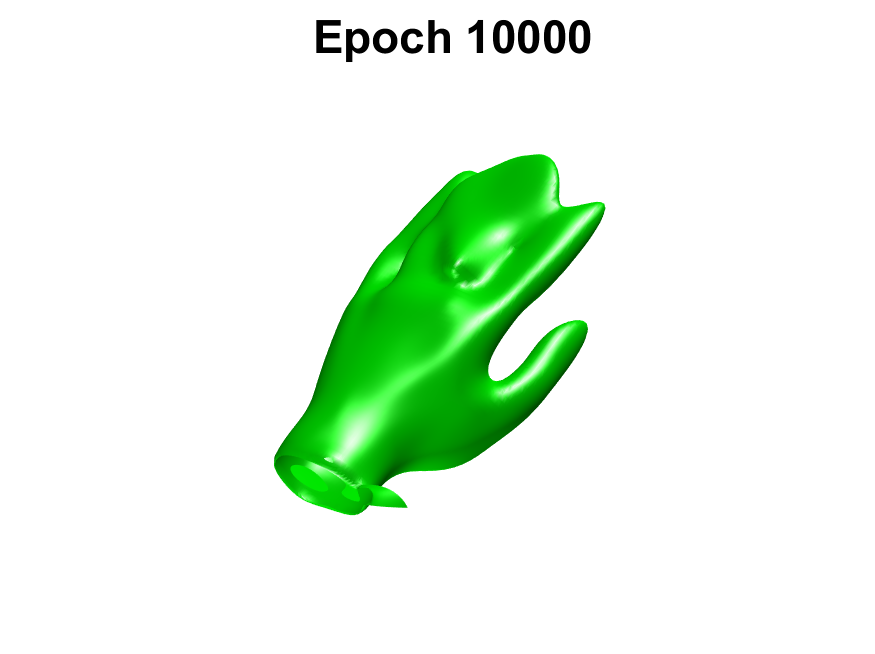}}
\subfigure[]{ \label{Ex_hand_resnet_iter21100}
\includegraphics[width=1.45in]{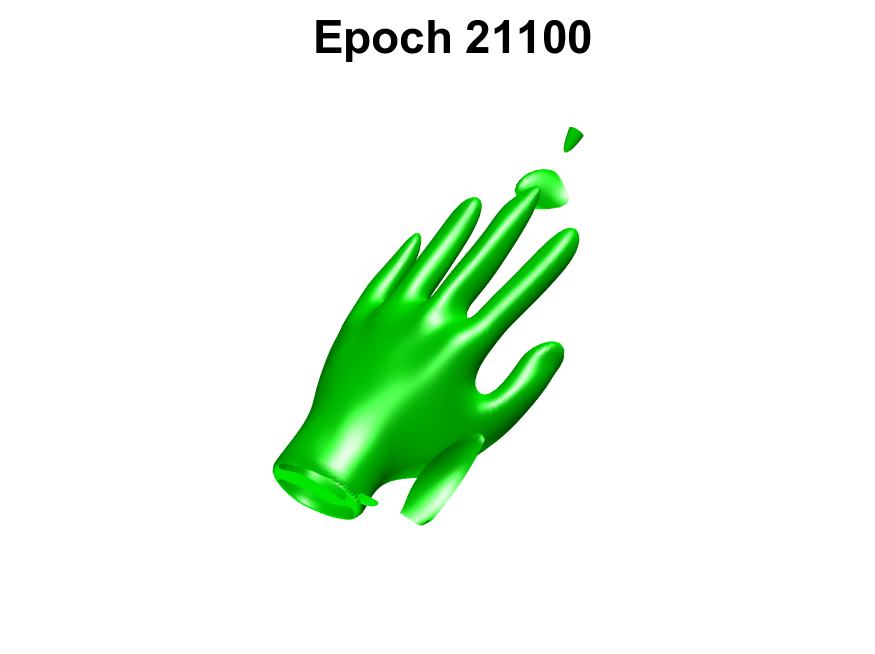}}
\subfigure[]{ \label{Ex_hand_sqr_iter50}
\includegraphics[width=1.45in]{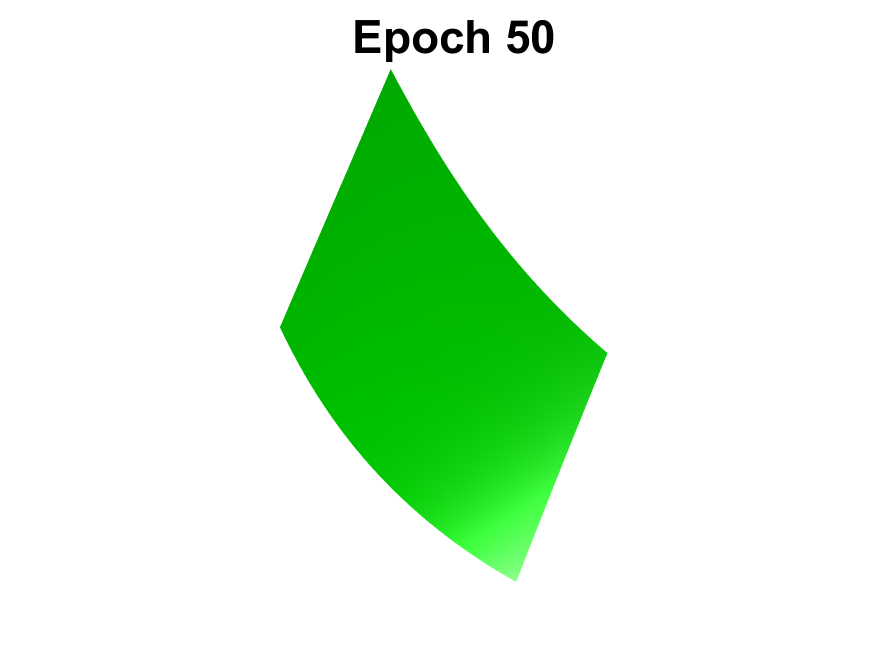}}
\subfigure[]{ \label{Ex_hand_sqr_iter5000}
\includegraphics[width=1.45in]{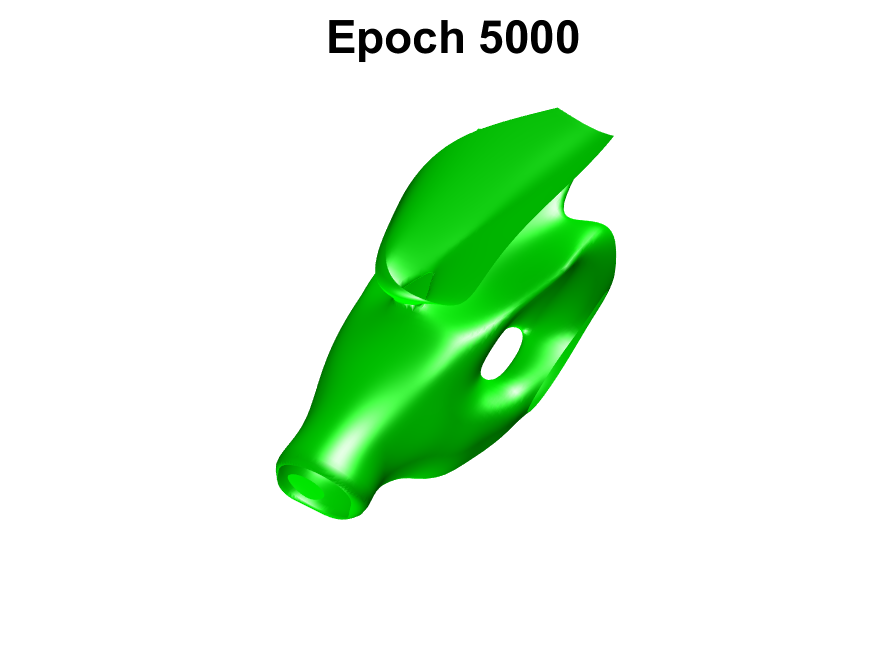}}
\subfigure[]{ \label{Ex_hand_sqr_iter10000}
\includegraphics[width=1.45in]{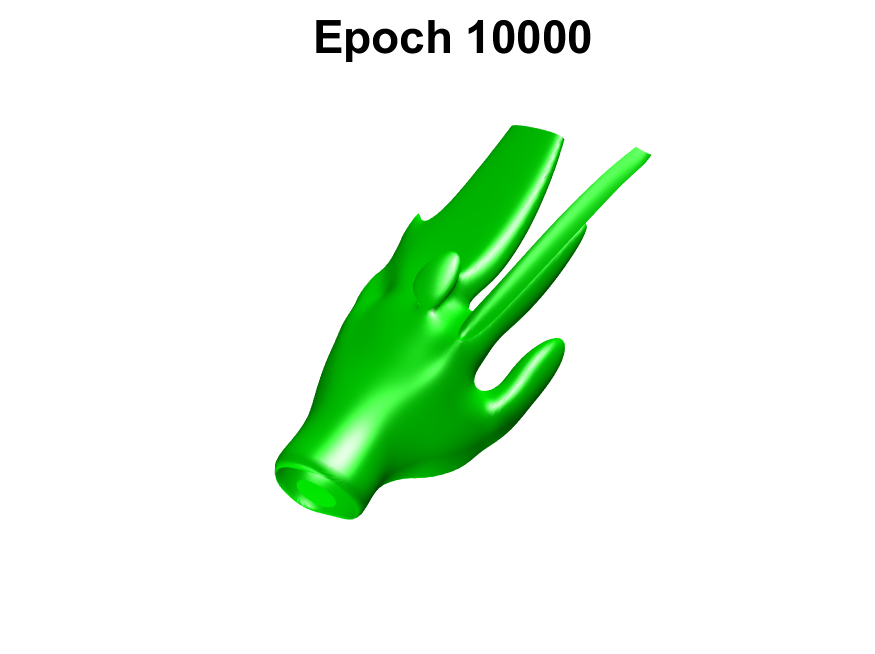}}
\subfigure[]{ \label{Ex_hand_sqr_iter18000}
\includegraphics[width=1.45in]{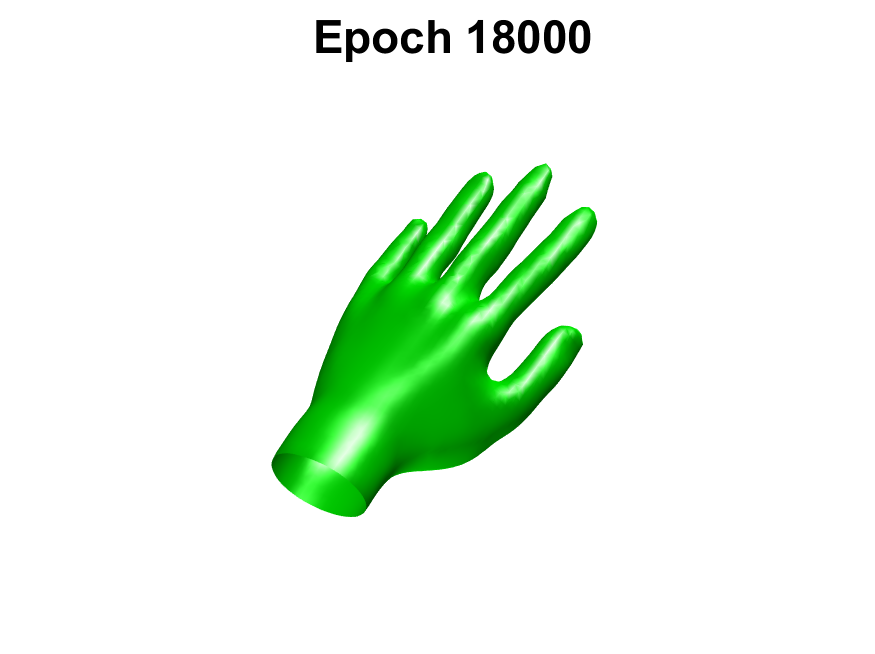}}
\caption{The profile of the simulated hand. The top panel: plain network,  middle panel:  Hw, bottom panel: SqrHw.  } \label{Ex1_2}
\end{figure}


\begin{figure}[!h]
\centering%
\subfigure[plain network]{ \label{normW_pnn}
\includegraphics[width=3.0in]{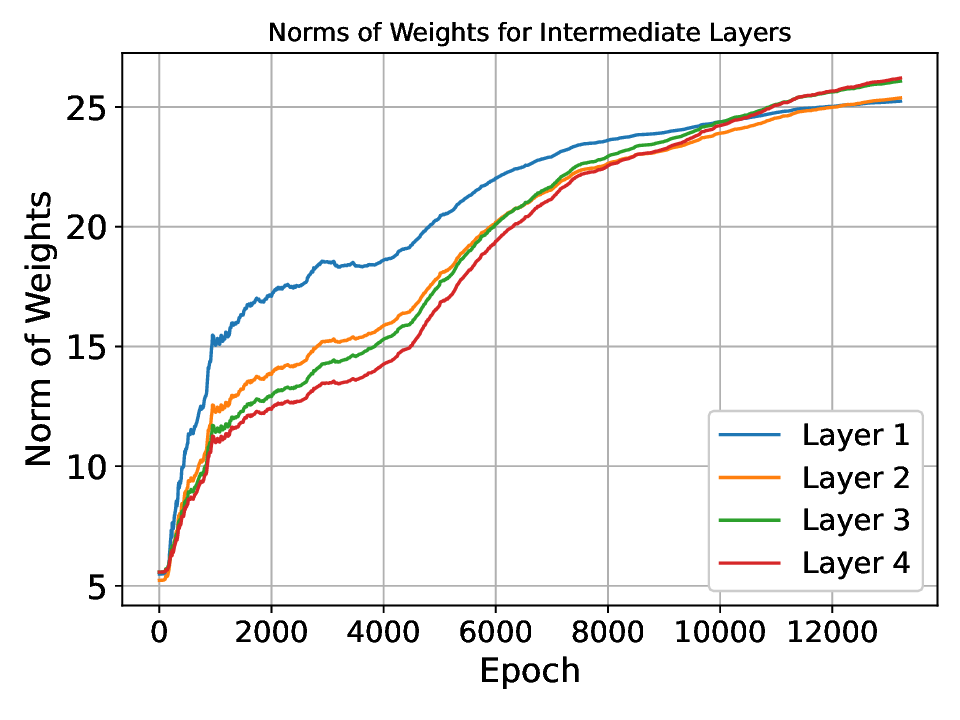}}
\subfigure[SqrHw]{ \label{normW_shw}
\includegraphics[width=3.0in]{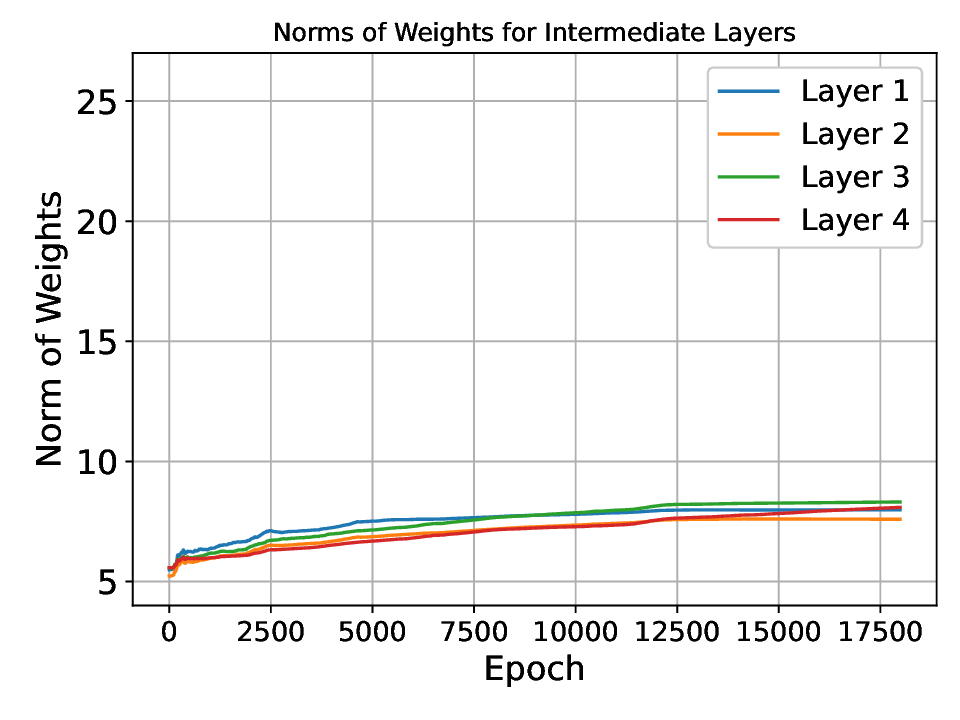}}
\caption{The profile of the norm of the weights with respect to the epochs. } \label{Ex1_4}
\end{figure}

To explore the factors underlying the disparities between the Pn and SqrHw models, we conducted an examination of the Frobenius norm of the weights updated across all hidden layers throughout every epoch. For this investigation, we selected the Frobenius norm, denoted as
\begin{equation}
 ||\textbf{W}||_F = \sqrt{\sum_{i=1}^{\mathcal{T}} \sum_{j=1}^{\mathcal{N}} |W_{ij}|^2}
\end{equation}
due to its ability to capture the overall magnitude and fluctuations of weight matrices. Here, $\mathcal{T}$ represents the number of epochs, and $\mathcal{N}$ indicates the number of weights across all layers at a specific epoch.
The resulting visualizations, presented in Fig.~\ref{Ex1_4}, reveal distinct patterns in the evolution of weight norms between the two models. Specifically, Fig.\ref{normW_pnn} depicts the fluctuating behavior of weight norms in the Pn model, while Fig.~\ref{normW_shw} illustrates a more consistent trend in the SqrHw model.

Notably, the SqrHw model demonstrates a convergence to stable weight norms, whereas the plain network continues to experience a significant increase in norms. This divergence in weight behavior can be attributed to the fact that the carry gate in SqrHw facilitates smoother optimization by providing clear paths for the flow of gradients \cite{Srivastava15}, allowing for easier weight updates. This smoother optimization process contributes to the observed stability in the evolution of weight norms in the SqrHw model.
Therefore, the introduction of carry gate in the SqrHw architecture plays a crucial role in addressing optimization challenges encountered in deep networks, leading to more stable weight norms and faster convergence compared to the Pn.

\begin{figure}[!h]
\centering%
\subfigure[Pn, $h=1$]{ \label{normG_pnn_l1}
\includegraphics[width=3.0in]{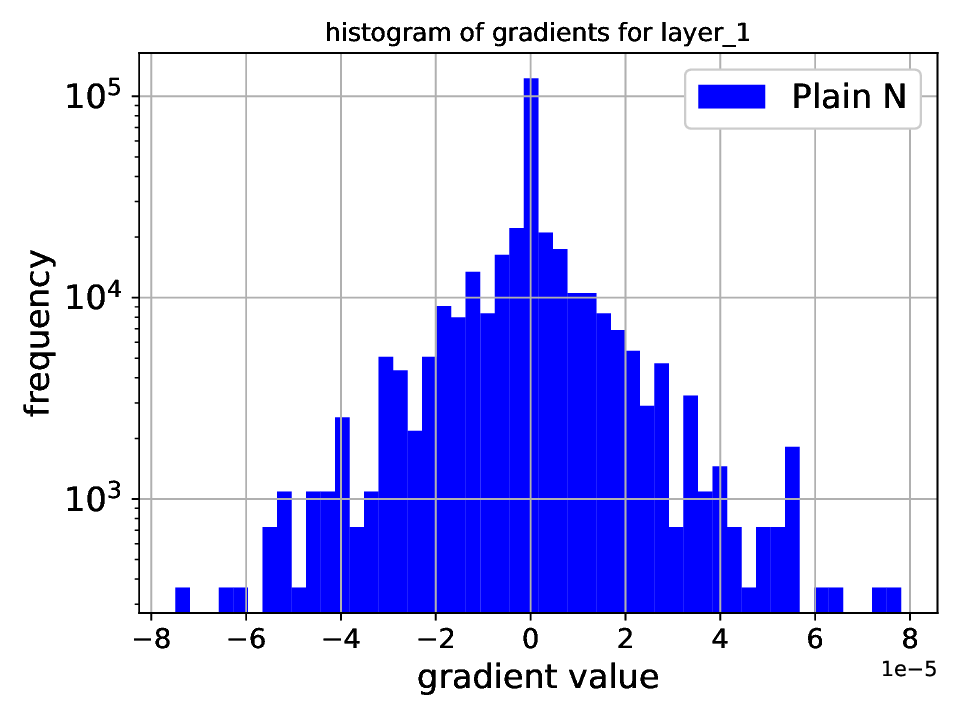}}
\subfigure[Pn, $h=4$]{ \label{normG_pnn_l4}
\includegraphics[width=3.0in]{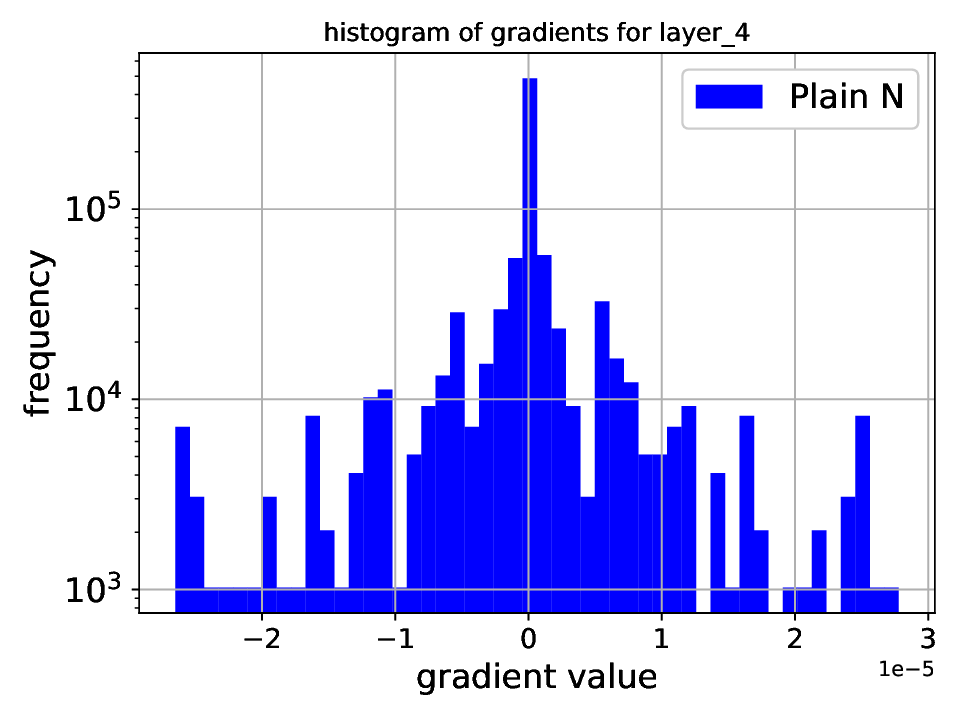}}
\subfigure[SqrHw, $h=1$]{ \label{normG_shw_l1}
\includegraphics[width=3.0in]{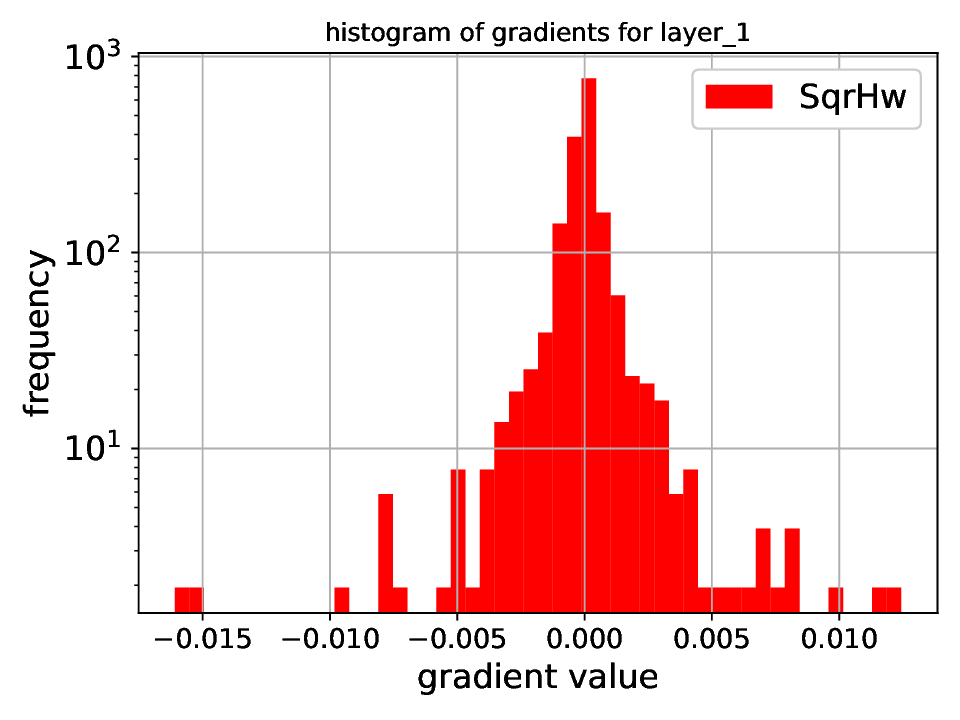}}
\subfigure[SqrHw, $h=4$]{ \label{normG_shw_l4}
\includegraphics[width=3.0in]{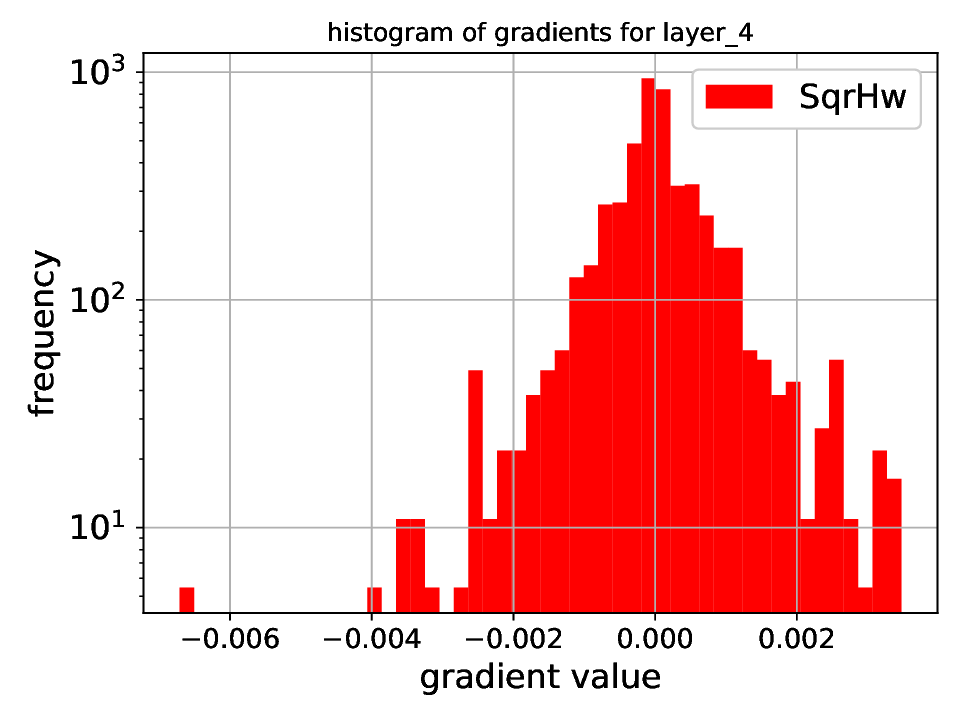}}
\caption{The profile of the histogram of back-propagated gradients with respect to the weight at epoch=12000 and different indices of the hidden layer $(h)$. } \label{Ex1_5}
\end{figure}

It is noticed that, the SqrHw model exhibits a trend towards stable weight norms, while the plain network continues to display a notable increase in norms over time. This divergence in weight behavior can be attributed to the inclusion of the carry gate in the SqrHw model, which facilitates smoother optimization by establishing clear pathways for the flow of gradients \cite{Srivastava15}. This smoother optimization process contributes to the observed stability in the evolution of weight norms in the SqrHw model.

Moreover, delving into the root cause of plain network's incapacity to provide precise predictions, we turn to the seminal works of Glorot and Bengio (2010) \cite{Glorot10} and Wang et al. (2021) \cite{Wang21}. We delve into the distribution of back-propagated gradients concerning the neural network weights throughout the training process, as illustrated in Fig. \ref{Ex1_5} for various hidden layers $h = 1$ and $4$. The top panel showcases outcomes for the plain network, while the bottom panel exhibits results for SqrHw.
Our investigation reveals a significant difference: the back-propagated gradients for SqrHw exhibit larger magnitudes compared to plain network, indicative of a resolution to the vanishing gradient issue. This phenomenon owes its existence to the presence of the carry gate within the Highway network, enabling gradients to circumvent multiple layers and thereby facilitating more efficient optimization. This innovation, initially proposed by He et al. (2015) \cite{Srivastava15}, translates into enhanced training stability, accelerated convergence, and ultimately, superior accuracy across the board.

 \begin{figure}[!h]
\centering%
\subfigure[$n_l=3$]{ \label{Ex_hand_nn100_nl3}
\includegraphics[width=1.45in]{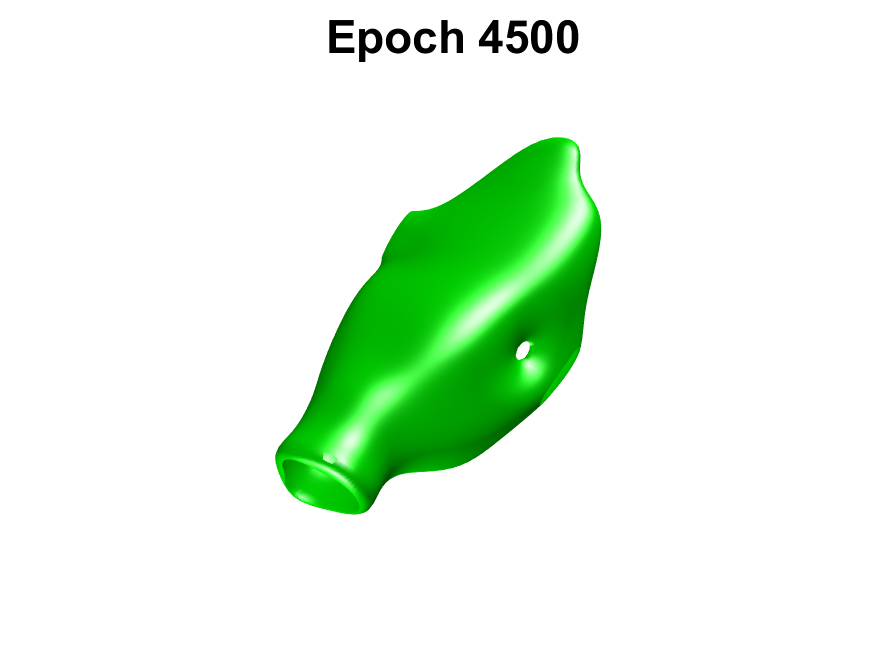}}
\subfigure[$n_l=5$]{ \label{Ex_hand_nn100_nl5_f}
\includegraphics[width=1.45in]{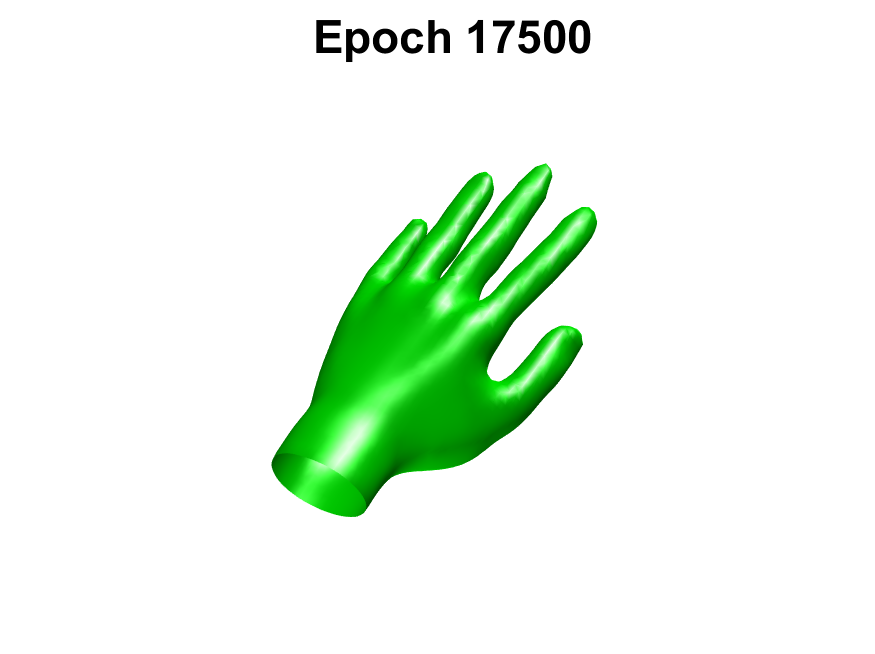}}
\subfigure[$n_l=10$]{ \label{Ex_hand_nn100_nl10_f}
\includegraphics[width=1.45in]{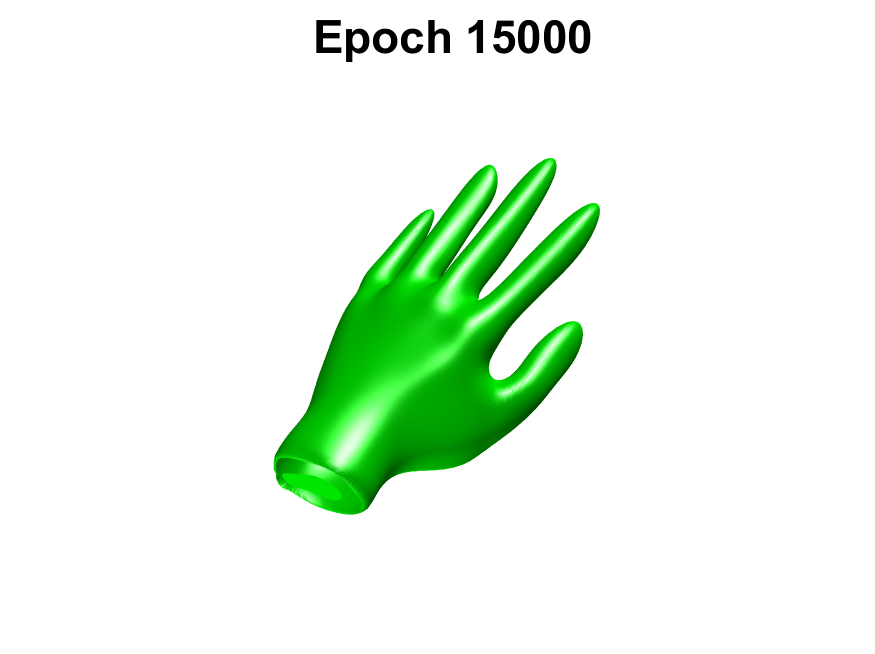}}
\subfigure[$n_l=15$]{ \label{Ex_hand_nn30_nl7}
\includegraphics[width=1.45in]{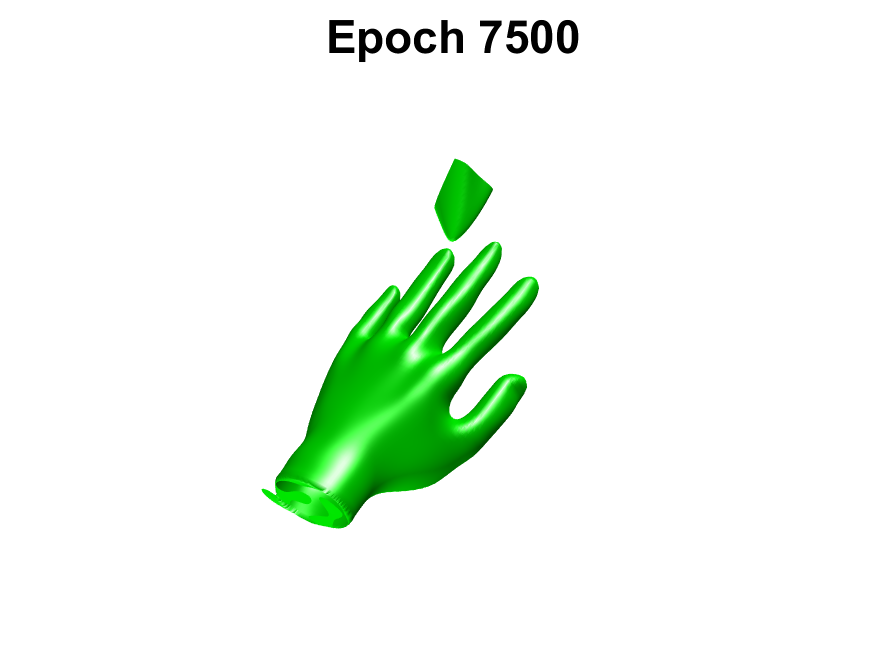}}
\caption{The profile of the simulated hand for various number of hidden layers.} \label{Ex1_3}
\end{figure}
 
Next, we investigate the impact of varying the number of hidden layers in for SqrHw network. Figure~\ref{Ex1_3} illustrates the results obtained when altering the number of hidden layers, ranging from 3 to 5, 10, and finally 15 hidden layers. These experiments were conducted with the same settings as the previous figure, with 100 neurons allocated for each hidden layer.

Several key observations can be made:

\begin{itemize}
  \item Utilizing only three hidden layers results in a hand shape that has not fully formed yet.
  \item Conversely, employing a larger number of hidden layers, such as 15, introduces additional surfaces.
  \item The configurations with 5 and 10 hidden layers appear to work well in generating accurate results.
  \item Throughout all numerical examples in this section, employing 5 hidden layers consistently produced favorable outcomes.
\end{itemize}

These findings emphasize the delicate balance between the number of hidden layers and the overall quality of the generated surfaces. The results underscore the importance of selecting an appropriate number of layers.

}
\end{example}

\begin{example}\rm{

\begin{figure}[!h]
\centering%
\subfigure[]{ \label{Ex_bodyUterus_uterusPts}
\includegraphics[width=2.5in]{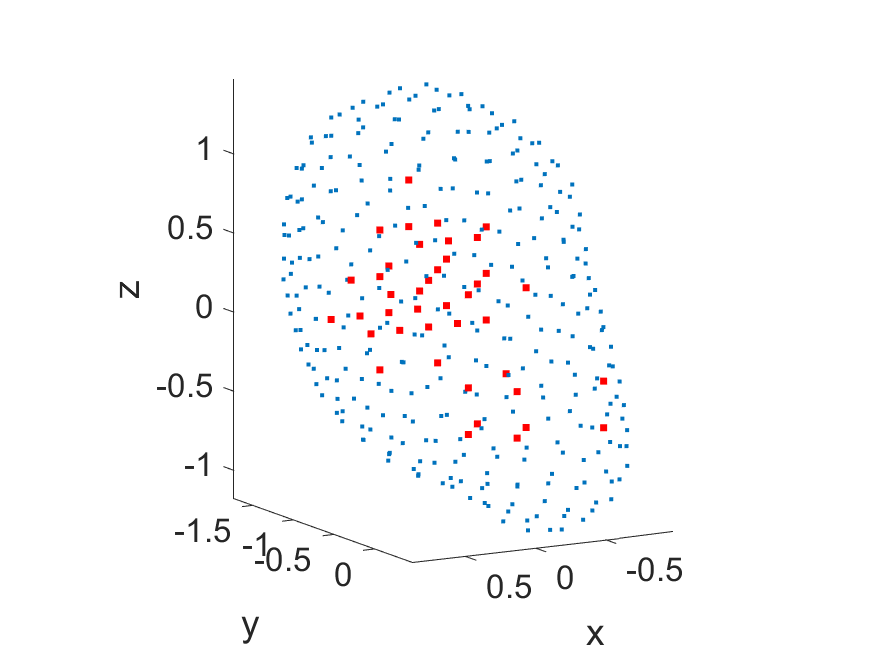}}
\subfigure[]{ \label{Ex_bodyUterus_bodyPts}
\includegraphics[width=2.5in]{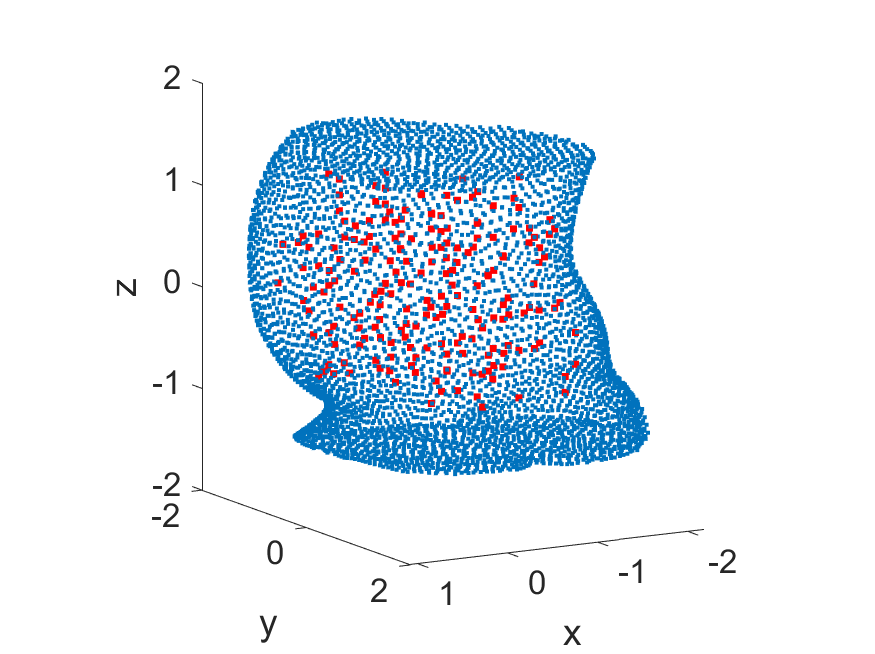}}
\caption{The profile of the point distribution in human (a) abdomen and (b) uterus. } \label{Ex2_1}
\end{figure}

\begin{figure}[!h]
\centering%
\subfigure[]{ \label{Ex_bodyUterus_BU}
\includegraphics[width=2.5in]{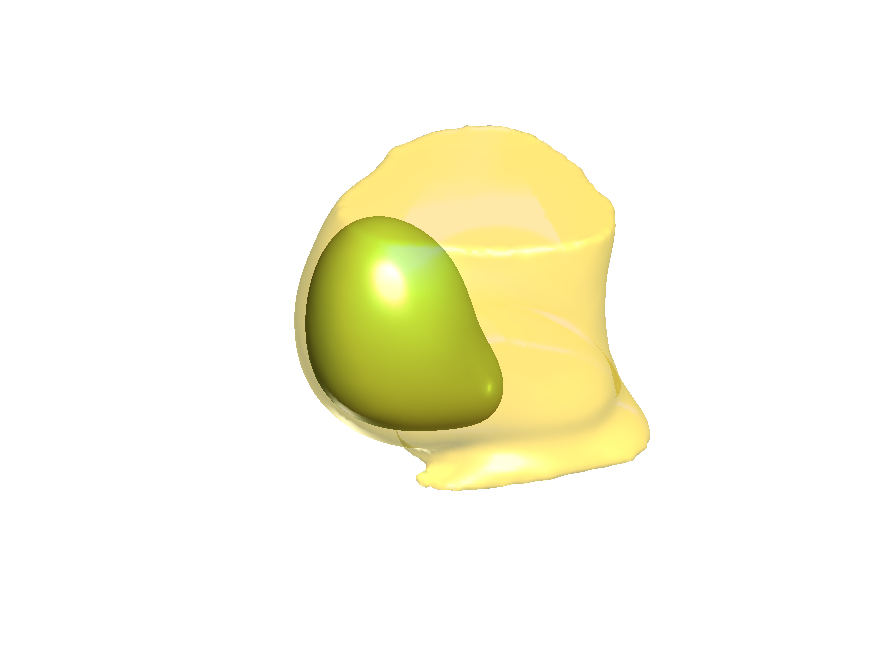}}
\subfigure[]{ \label{Ex_bodyUterus_BU_rear}
\includegraphics[width=2.5in]{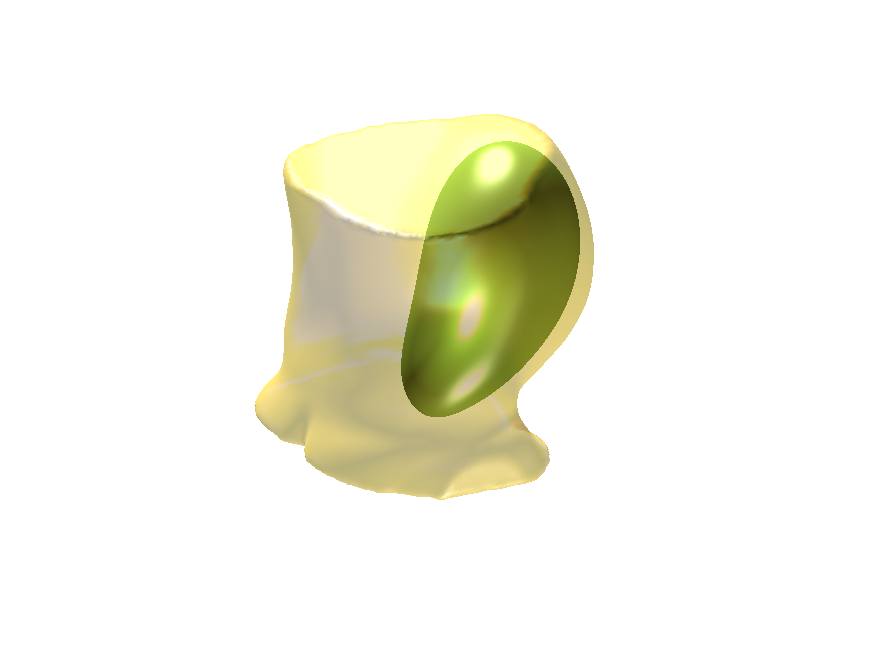}}
\caption{The profile of the simulated body and uterus. } \label{Ex2_2}
\end{figure}

In this instance, we embark on the endeavor of reconstructing the three-dimensional surface of the pregnant human uterus and abdomen  where the point distribution are represented in Fig.~\ref{Ex_bodyUterus_uterusPts} and Fig.~\ref{Ex_bodyUterus_bodyPts}, respectively.

The precise reconstruction of the uterus-abdomen geometry carries significant importance, particularly in maintaining the accuracy of electro-myometrial imaging (EMMI). EMMI is an innovative electro-physiology imaging modality designed to non-invasively capture the electrical activation patterns of the uterus as it undergoes mechanical contraction. To gather the data points for the abdomen, we employed an optical scanning device. For the uterus, data points were obtained through Magnetic Resonance Imaging (MRI). It is worth noting that MRI is a costly procedure and can be conducted exclusively in clinics equipped with MRI facilities.  All data were graciously provided by the Integrated Biomedical Imaging Laboratory at the Department of Obstetrics and Gynecology, School of Medicine, Washington University at St. Louis.

\begin{figure}[!h]
\centering%
\subfigure[Data]{ \label{Ex_bodyUterus_uterusPts_miising}
\includegraphics[width=1.45in]{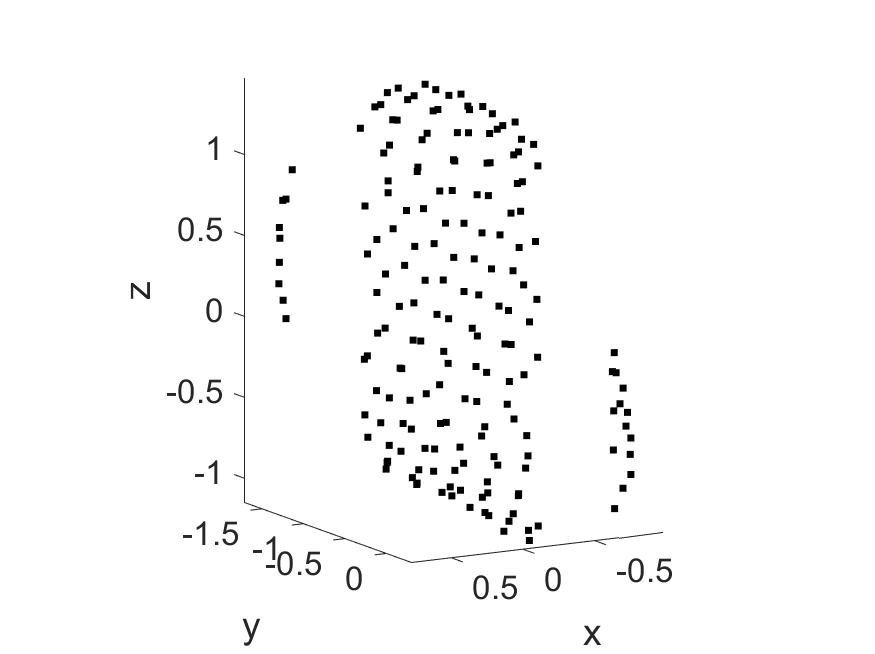}}
\subfigure[plain network]{ \label{Ex_bodyUterus_missing_pn}
\includegraphics[width=1.45in]{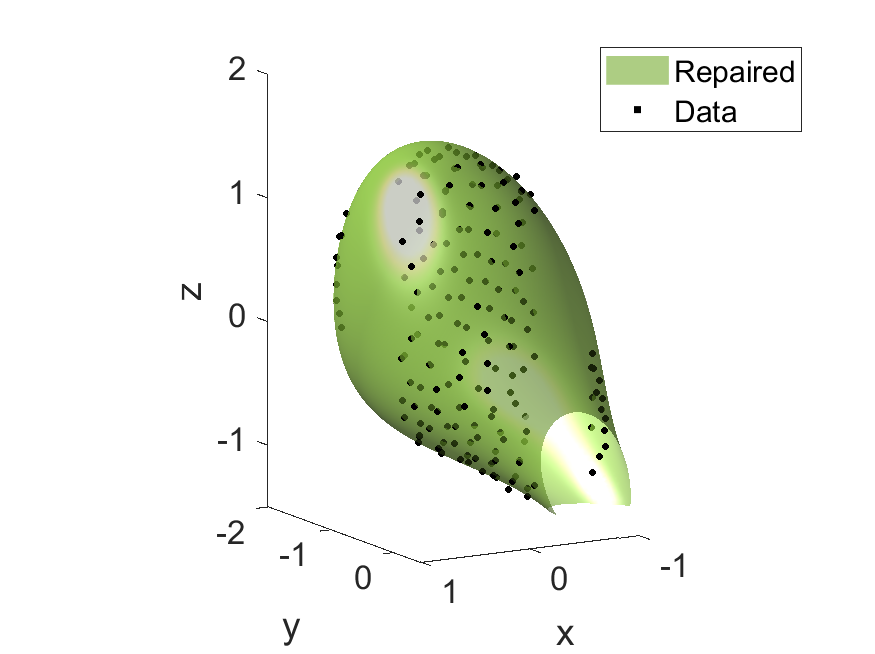}}
\subfigure[Hw]{ \label{Ex_bodyUterus_missing_resnet}
\includegraphics[width=1.45in]{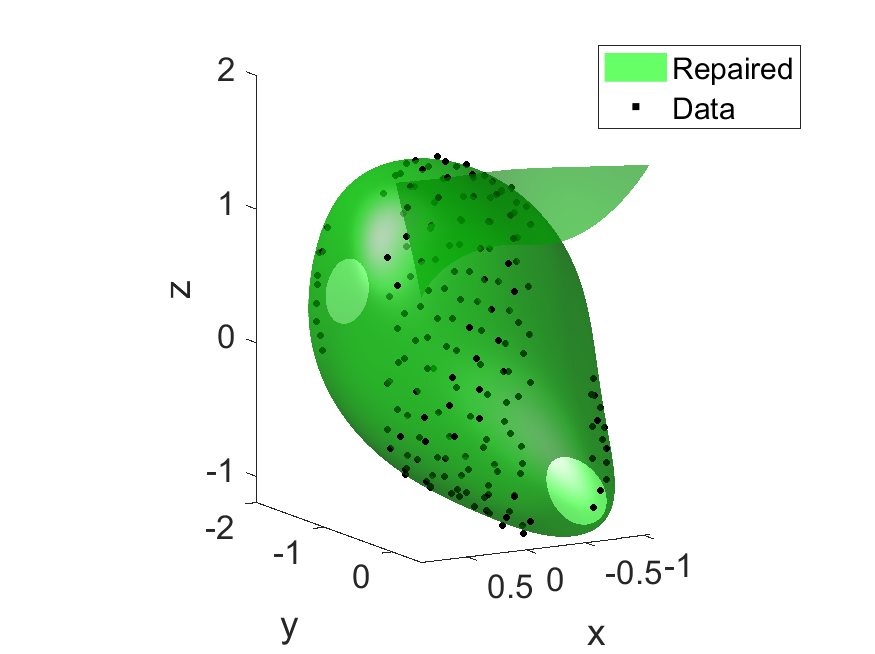}}
\subfigure[SqrHw]{ \label{Ex_bodyUterus_missing_compa_pts}
\includegraphics[width=1.45in]{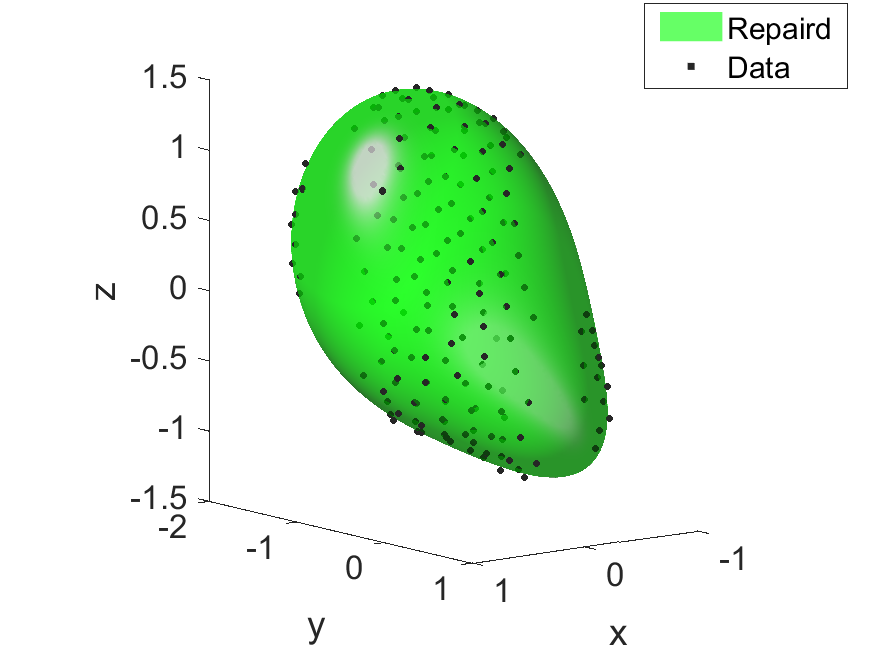}}
\caption{The profile of the (a) point distribution on the surface of uterus with missing data (b) Repaired surface using plain network, and (c) Repaired surface using Hw, and (d) Repaired surface using SqrHw. } \label{Ex2_3}
\end{figure}

For this example, we utilized 5000 data points for reconstructing the abdominal surface and 300 points for the uterus. The network consists of five hidden layers, each with 50 neurons.
The final results, depicted in Fig.~\ref{Ex2_2}, showcase impressively accurate reconstructions from various perspectives. We also explored a scenario involving missing data points, specifically in the context of uterine surface reconstruction, as illustrated in Fig.~\ref{Ex2_3}. This scenario entailed two distinct areas of missing data: one in the upper region and another in the lower part. In Fig.~\ref{Ex_bodyUterus_uterusPts_miising}, the available data points are highlighted, totaling 182 points, while it is noteworthy that 40\% of the data is missing, a noticeable contrast compared to Fig.~\ref{Ex_bodyUterus_uterusPts}. We retained the same set of interior points as in the previous case.
Subsequently, Fig.~\ref{Ex_bodyUterus_missing_pn} reveals that the Plan N algorithm effectively predicted the surface in the upper region with missing data. However, in the lower region where data was absent, the algorithm struggled to reconstruct the surface. Furthermore, Fig.~\ref{Ex_bodyUterus_missing_resnet} demonstrates the uterine surface reconstruction using the Hw network. Similar to the plain network results, a non-contiguous surface is observed in the region with missing data in the lower part, along with the appearance of an additional surface.
On the other hand, the utilization of the SqrHw yielded remarkable results, effectively predicting the surface over the missing data points, highlighting its robust capabilities in handling such challenges.

}
\end{example}

\begin{example}\rm{
\begin{figure}[!h]
\centering%
\includegraphics[width=2.05in]{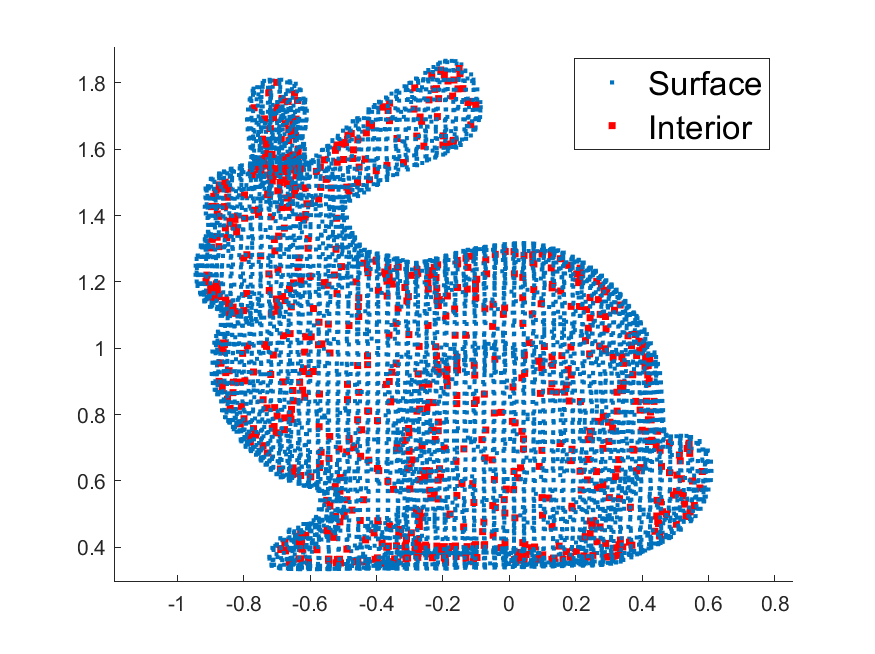}
\caption{The profile of the  point distribution for the Stanford bunny.  } \label{Ex_bunny_pts}
\end{figure}
In our final example, we tackle the reconstruction of a complex domain, the Stanford Bunny \cite{bunny} as the point distribution is shown in  Fig.~\ref{Ex_bunny_pts}. 

\begin{figure}[!h]
\centering%
\subfigure[]{ \label{Ex_bunny_pn_front}
\includegraphics[width=1.55in]{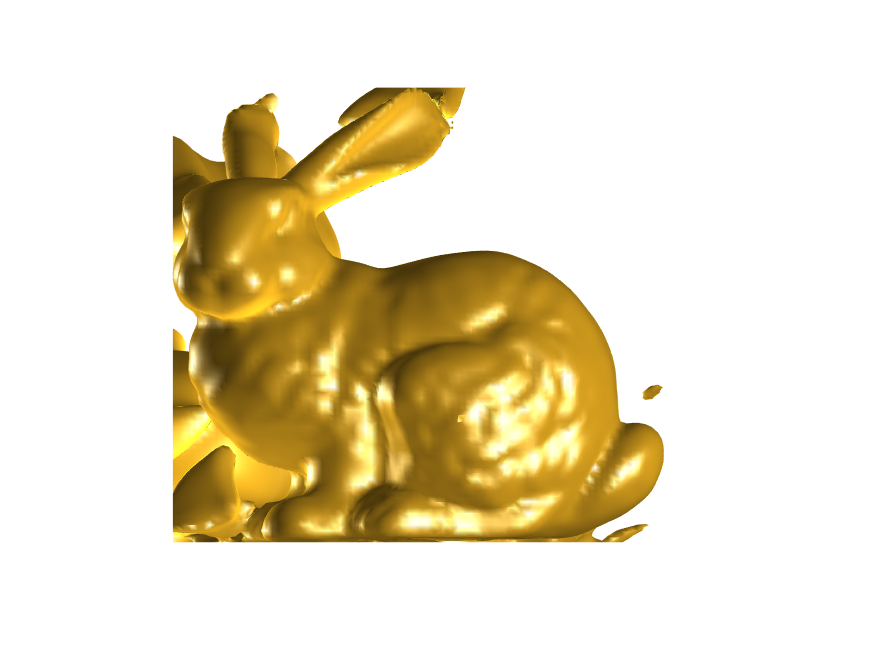}}
\subfigure[]{ \label{Ex_bunny_resnet_front}
\includegraphics[width=1.55in]{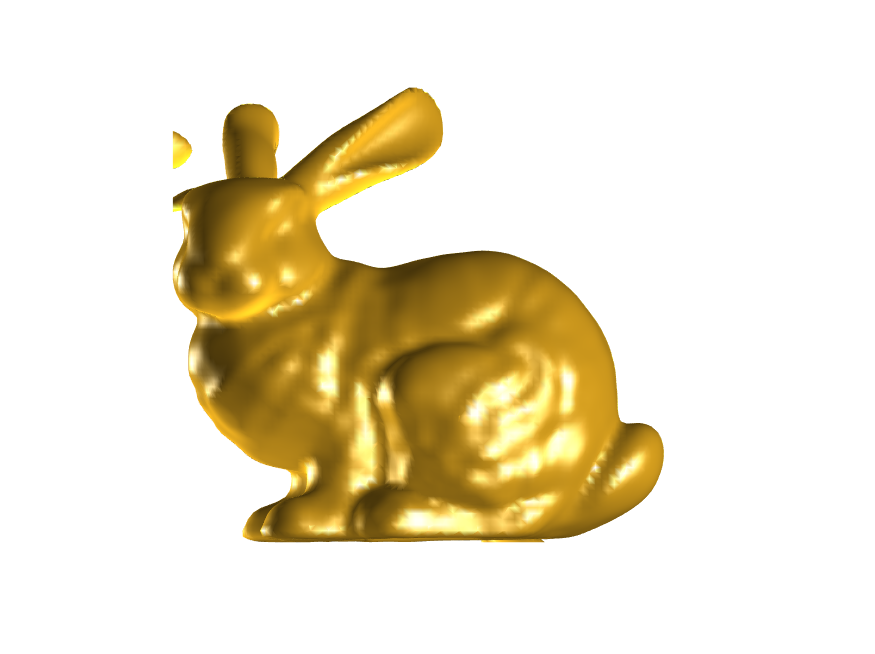}}
\subfigure[]{ \label{Ex_bunny_sqr_front}
\includegraphics[width=1.55in]{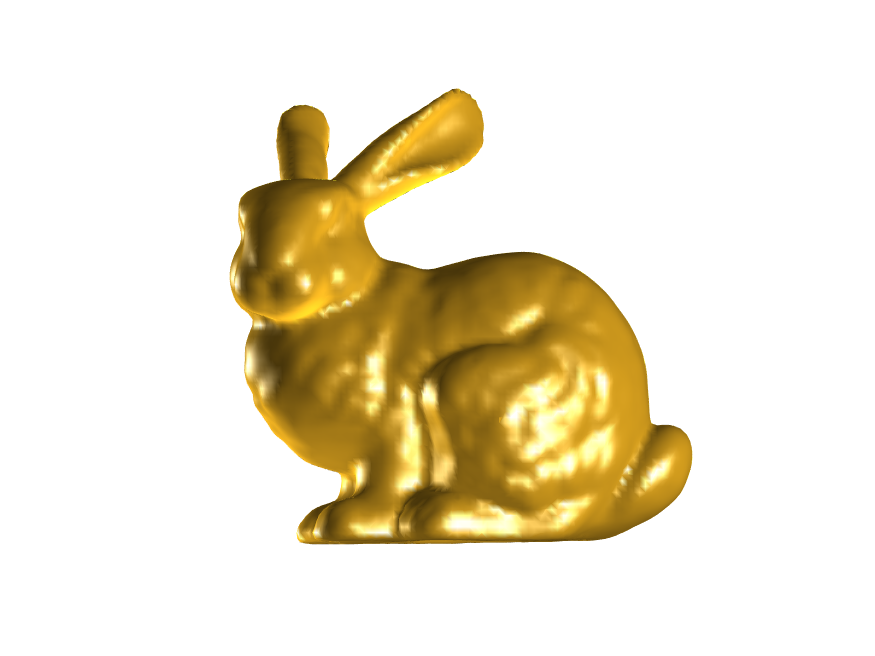}}
\subfigure[]{ \label{Ex_bunny_pn_rear}
\includegraphics[width=1.55in]{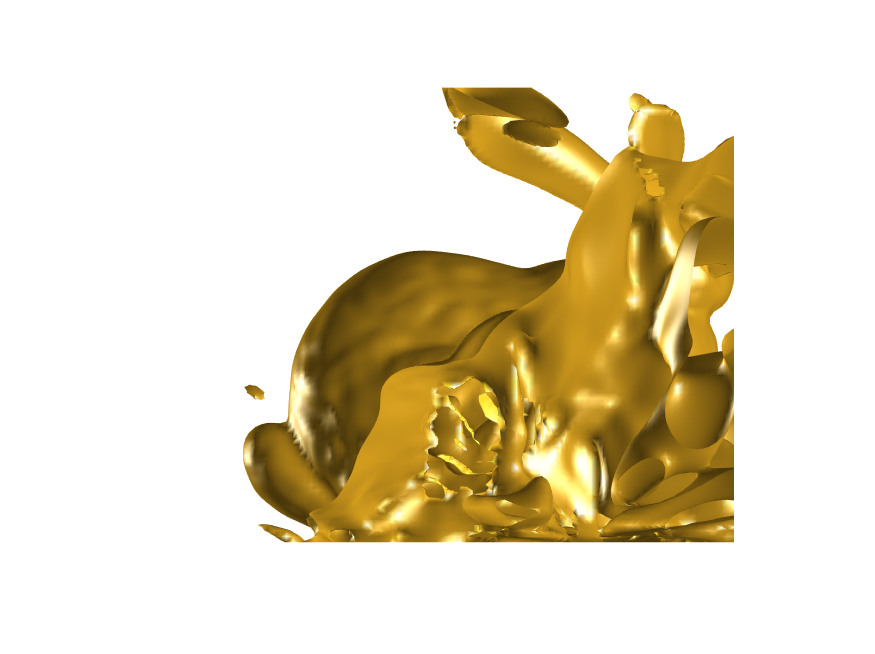}}
\subfigure[]{ \label{Ex_bunny_resnet_rear}
\includegraphics[width=1.55in]{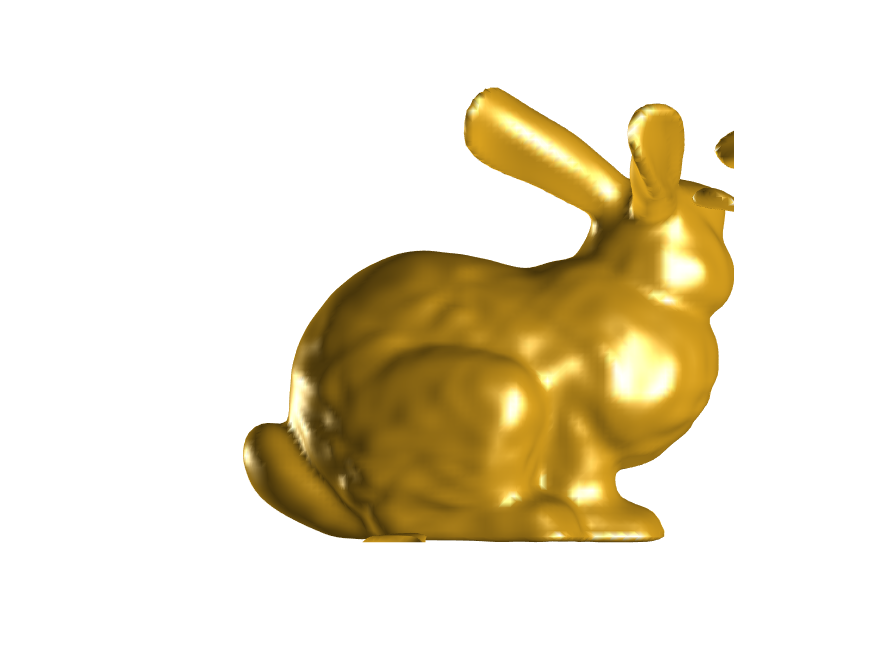}}
\subfigure[]{ \label{Ex_bunny_sqr_rear}
\includegraphics[width=1.55in]{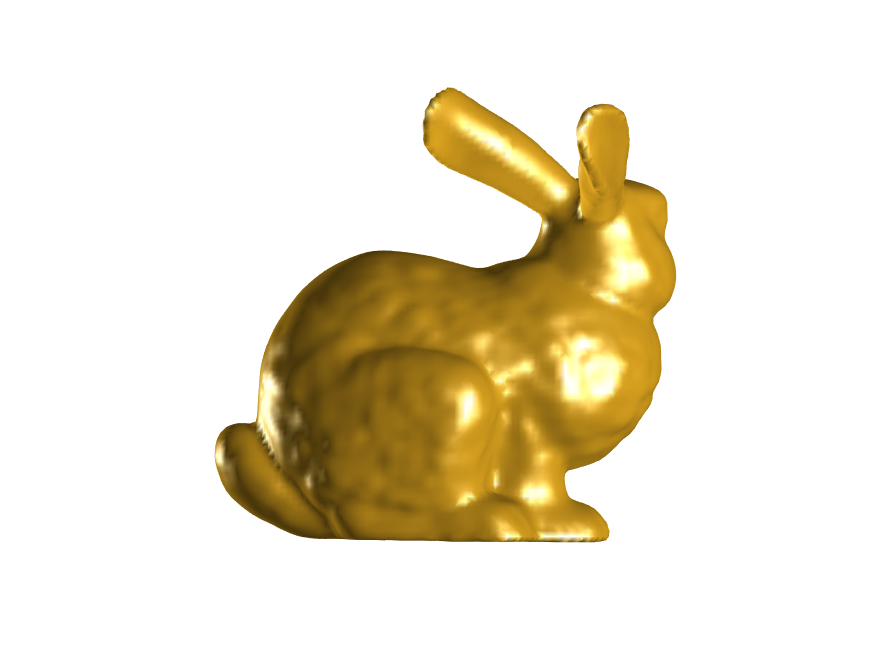}}
\caption{The surface of the simulated Stanford bunny. The left panel: plain network,  middle panel:  Hw, and right panel: SqrHw. } \label{Ex3_2}
\end{figure}

\begin{figure}[!h]
\centering%
\subfigure[]{ \label{Ex_bunny_pn_front_ext}
\includegraphics[width=1.45in]{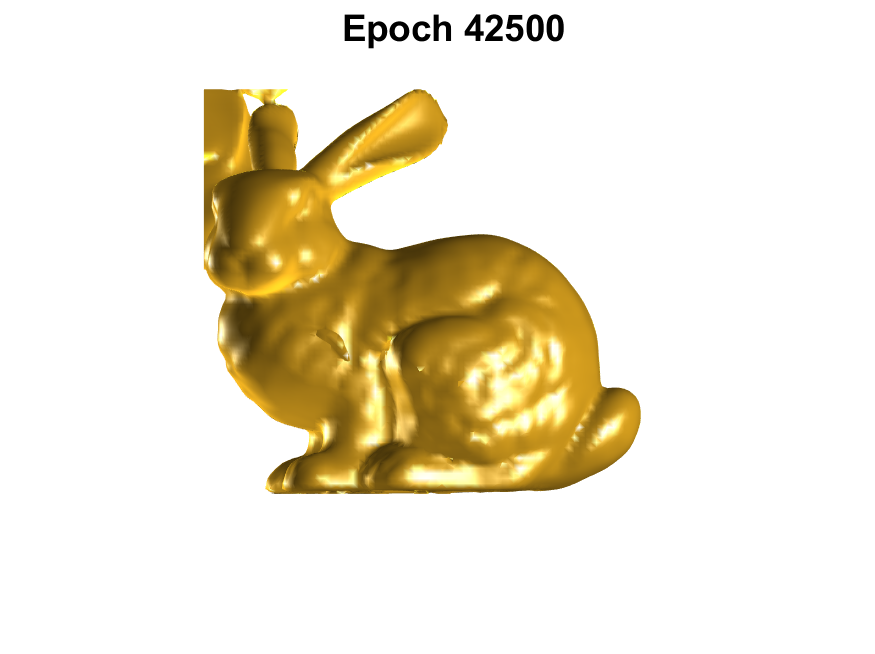}}
\subfigure[]{ \label{Ex_bunny_pn_rear_ext}
\includegraphics[width=1.45in]{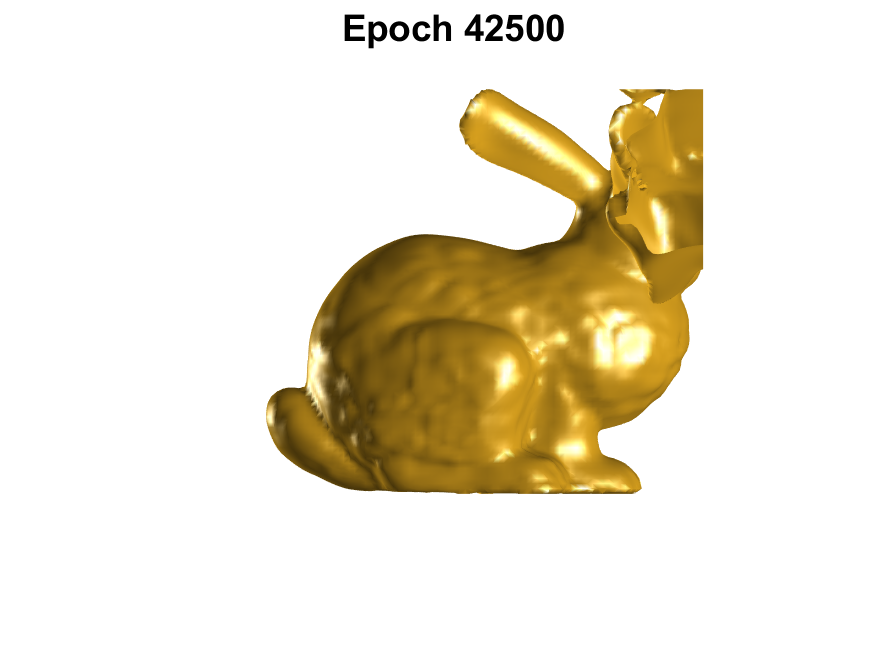}}
\subfigure[]{ \label{Ex_bunny_resnet_front_ext}
\includegraphics[width=1.45in]{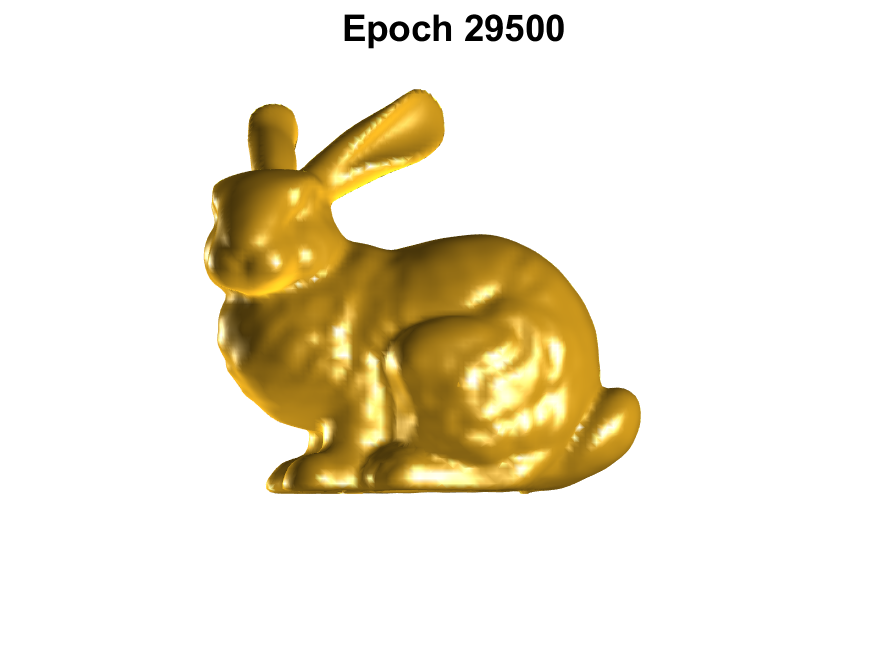}}
\subfigure[]{ \label{Ex_bunny_resnet_rear_ext}
\includegraphics[width=1.45in]{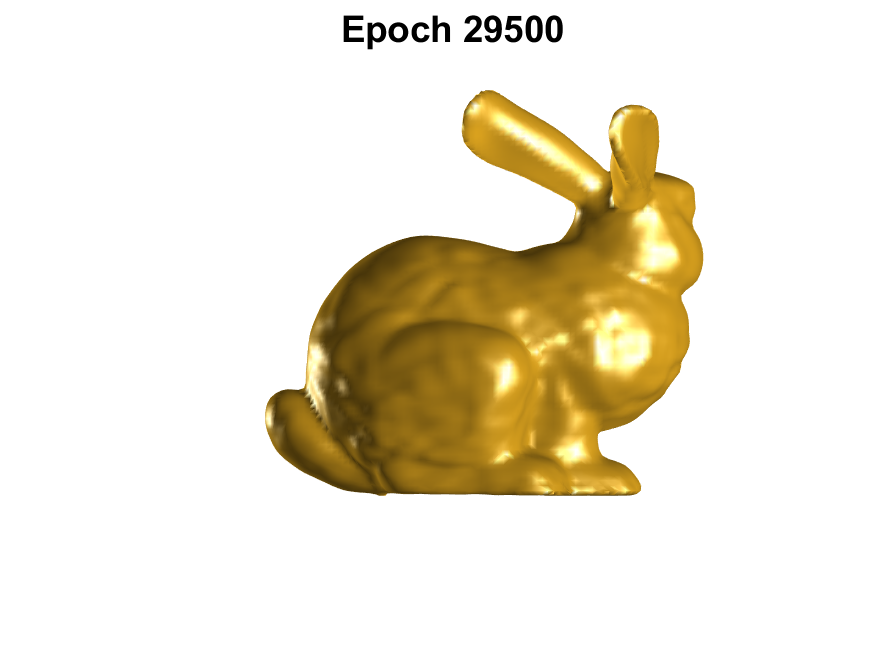}}
\caption{The improved surface visualization of the simulated Stanford bunny utilizing exterior points. The first two images on the left depict results from the plain network method, while the subsequent two images showcase outcomes from the Hw network, with $n_i=1000$ and $n_e=1000$. } \label{Ex3_3}
\end{figure}

Figure~\ref{Ex3_2} provides the outcomes when applying the plain network (left panel), Hw (middle panel), and the SqrHw (right panel). The top row showcases the bunny's front view, while the bottom row reveals the rear view. For this illustration, we maintain a consistent set of parameters, $n_s=8000$ and $n_i=3000$. We also implemented a neural network with five hidden layers, each containing 50 neurons.

Notably, the plain network results in a relatively unclear representation of the bunny, particularly when viewed from the rear perspective (see Fig.~\ref{Ex_bunny_pn_rear}). In contrast, both simplified Hw and SqrHw deliver more accurate representations. However, as previously observed, the utilization of simplified Hw introduces some additional surface features.

To improve our results, we introduced 1000 exterior points $(n_e)$ and reran the simulation using both plain network and the simplified Hw methods. The outcomes are illustrated in Fig.~\ref{Ex3_3}, where the first two images display the plain network results, and the subsequent two images showcase the results obtained using the Hw approach. It is evident that the results obtained using plain network demonstrate a significant enhancement compared to those shown in Fig.~\ref{Ex_bunny_pn_front} and Fig.~\ref{Ex_bunny_pn_rear}.
Moreover, the Hw results also display a significant refinement in surface construction when exterior points are introduced. This underlines the substantial impact of exterior points on the quality of reconstruction.

}
\end{example}

\section{Conclusion}
Our study provides a thorough analysis of neural network architectures for surface reconstruction from point clouds within the realm of computer graphics. Through empirical evaluations and numerical analyses, we have shed light on the efficacy of different network structures in tackling the challenges inherent in surface reconstruction tasks, thereby advancing the understanding of deep learning methods for surface generation.

\begin{itemize}
\item Our proposed Squared Highway (SqrHw) network architecture exhibits superior performance compared to conventional plain neural networks.
\item The SqrHw network demonstrates faster convergence and more stable optimization behavior, thanks to the introduction of a carry gate in the architecture. This architectural feature facilitates efficient gradient flow and mitigates the vanishing gradient problem.
\item Empirical evaluations on various datasets, encompassing simple geometric shapes and complex objects like the human hand and the Satnford Bunny, underscore the effectiveness of the proposed architecture in accurately reconstructing surfaces from point clouds.
\item Analysis of network parameters, such as the number of hidden layers and neurons, emphasizes the significance of selecting an appropriate architecture configuration to attain optimal reconstruction results.
\end{itemize}
\noindent
The analysis of the Frobenius norm of weight updates revealed distinct patterns in weight evolution between the two models. Specifically, the SqrHw model demonstrated a convergence to stable weight norms, indicating smoother optimization compared to the fluctuating and increasing behavior observed in the Pn model.
Furthermore, examination of back-propagated gradients highlighted the SqrHw's ability to mitigate the vanishing gradient issue, leading to enhanced training stability and faster convergence. The presence of the carry gate in SqrHw facilitates more efficient optimization by enabling gradients to bypass multiple layers, thereby addressing optimization challenges encountered in deep networks.

\noindent
Future research can extend the application of the MLPs architecture to various purposes, including physics-informed neural networks. Additionally, one can incorporate partial differential equations such as the Helmholtz equations for computer graphics problems \cite{Zheng20} within the context of physics-informed neural networks.

\section*{Acknowledgements}
The authors gratefully acknowledge the financial support of the National Science and Technology Council of Taiwan under grant numbers 112-2221-E-002-097-MY3, 112-2811-E-002-020-MY3. We also want to acknowledge the NTUCE-NCREE Joint Artificial Intelligence Research Center and the National Center of High-performance Computing (NCHC) in Taiwan for providing computational and storage resources.

\end{document}